\RequirePackage{fix-cm}
\documentclass{article}
\PassOptionsToPackage{numbers, compress}{natbib}

\usepackage[main, final]{neurips_2026}

\usepackage[utf8]{inputenc}
\usepackage[T1]{fontenc}
\usepackage[hidelinks]{hyperref}
\usepackage{url}
\usepackage{graphicx}
\usepackage{amsmath}
\usepackage{amsfonts}
\usepackage{amssymb}
\usepackage{nicefrac}
\usepackage{microtype}
\usepackage{xcolor}
\usepackage{threeparttable}
\usepackage{tabularx}
\usepackage{booktabs}
\usepackage{multirow}
\usepackage{colortbl}
\usepackage{hhline}
\usepackage{subcaption}
\usepackage{caption}
\usepackage{arydshln}
\usepackage{wrapfig}
\usepackage{fontawesome}

\title{EchoKV: Efficient KV Cache Compression via Similarity-Based Reconstruction}

\author{%
  Shiyu Ji\thanks{Equal contribution.}\quad
  Yixuan Wang\footnotemark[1]\quad
  Yijun Liu\quad
  Qingfu Zhu\quad
  Wanxiang Che\thanks{Corresponding author.} \\
  Research Center for Social Computing and Interactive Robotics, \\
  Harbin Institute of Technology, China \\
  \texttt{\{syji,car\}@ir.hit.edu.cn} \\
  \faGithubAlt~\url{https://github.com/noforit/EchoKV}
}

\begin{document}
\maketitle

\begin{abstract}
The increasing memory demand of the Key-Value (KV) cache poses a significant bottleneck for Large Language Models (LLMs) in long-context applications.
Existing low-rank KV compression methods reduce this footprint by modifying model projections, limiting the flexibility to switch back to standard full-cache inference when sufficient memory is available.
In this paper, we propose EchoKV, a flexible KV cache compression framework that supports on-demand transitions from full KV caching to compressed caching. 
Unlike traditional compression-decompression paradigms, EchoKV utilizes a lightweight network to reconstruct the discarded KV components from a partial subset, exploiting intrinsic inter-layer and intra-layer similarities among attention heads. 
We further introduce a lightweight two-stage fine-tuning strategy, requiring only a few minutes on a single A100 GPU for Llama-3.1-8B.
Experimental results on LongBench and RULER demonstrate that EchoKV offers a competitive quality--cache trade-off across multiple compression ratios and backbone models while preserving the throughput of full-cache inference in short-context scenarios.

\end{abstract}

\section{Introduction}
While the KV cache significantly accelerates Large Language Model inference by reducing redundant computations,
it introduces a heavy memory burden that scales linearly with context length.
As LLMs advance to support massive context windows for applications like repository-scale code generation \citep{jimenez2023swe} and deep reasoning \citep{guo2025deepseek} via ultra-long Chain-of-Thought \citep{chen2025towards},
the memory overhead of the KV cache has become a prohibitive bottleneck,
catalyzing a surge of research \citep{shi2024keep} into compression methodologies.

Recent research on KV cache compression has primarily centered on low-rank sharing.
Similar to the Multi-Head Latent Attention (MLA) \citep{liu2024deepseek},
Palu \citep{chang2024palu} achieves training-free low-rank sharing by performing Singular Value Decomposition (SVD) within attention head groups, 
effectively reducing the KV cache footprint.
Furthermore, CommonKV \citep{wang2025commonkv} leverages the intrinsic similarity of KV caches across layers to extend low-rank sharing to a multi-head, multi-layer scope,
achieving superior performance under low memory budgets.

\begin{figure}[t]
    \centering
    \includegraphics[width=0.9\linewidth]{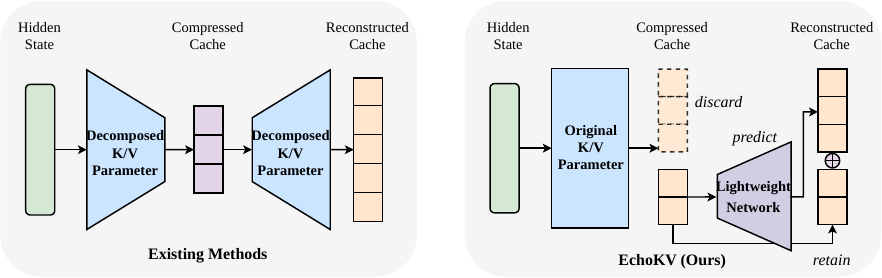}
    \caption{Illustration of the differences between existing low-rank sharing approaches and EchoKV.
Unlike the compression-decompression paradigm, EchoKV employs a lightweight network to reconstruct the discarded KV components from retained ones.
}
    \label{fig:intro}
\end{figure}

While these methods effectively compress the KV cache footprint during inference, they all involve modifying model parameters.
This parameter modification makes it difficult to switch between standard full-cache inference and compressed inference at runtime, thereby compromising inference performance when memory is abundant.
Although some online compression methods \citep{liu2024minicache,chang2025xkv} offer selective compression, 
they struggle to maintain performance under tight memory budgets
and incur additional latency overheads due to online compression.
Consequently, there remains a lack of a flexible compression approach capable of reconciling
\textbf{high performance on short contexts} with
\textbf{low resource costs on long contexts}.

In this paper, we propose EchoKV,
a more flexible KV cache compression scheme that enables on-demand switching from standard inference to compressed inference while maintaining performance.
As shown in Figure \ref{fig:intro},
unlike other low-rank methods that perform uniform compression and reconstruction on the entire KV cache,
EchoKV employs a lightweight network to predict the remaining KV pairs from a partial subset,
thereby enabling a seamless transition from standard attention.
Specifically, inspired by the observation of inter-layer \citep{lee2024infinigen,wang2025commonkv} and intra-layer similarities \citep{wang2025proxyattn,peng2025accelerating} in KV representations,
we propose to bypass the explicit compression process.
Instead, we directly train a lightweight network to predict the remaining KV cache based on partial KV information.
To ensure both the integrity and accuracy of the input information, 
we utilize the full KV cache from specific layers alongside a partial subset from the current layer as input features to predict the remaining KV components of the current layer.
Furthermore, we employ a memory-friendly two-stage fine-tuning strategy that enables rapid and low-cost acquisition of the lightweight reconstruction network, requiring only a few minutes on a single A100 GPU for Llama-3.1-8B.

To validate the effectiveness of the proposed method,
we evaluate models of different scales, ranging from 7B to 24B parameters, on two long-context benchmarks: LongBench \citep{bai2024longbench}, which comprises real-world data,
and RULER \citep{hsieh2024ruler}, which consists of synthetic data.
Experimental results demonstrate that EchoKV achieves competitive performance across different compression ratios, with particular advantages at lower cache budgets.
Moreover, the capability to flexibly switch from FullKV to EchoKV maintains high throughput for short inputs 
while keeping long-context KV cache within memory limits, 
which provides significant practical advantages relative to alternative techniques.
Through combination with low-rank techniques,
EchoKV yields additional performance gains, 
specifically tailored to the distinct properties of keys and values.

The main contributions of this paper are summarized as follows:
\begin{itemize}
    \item We propose EchoKV, a KV cache compression method that reconstructs the remaining cache using specific KV information. It enables flexible switching from FullKV to EchoKV.
    \item We introduce an efficient training method for the lightweight reconstruction network, allowing training to be completed in a time comparable to offline SVD.
    \item Experiments demonstrate that our proposed method achieves competitive performance across multiple benchmarks, compression ratios, and model scales while maintaining high throughput in short-context scenarios.
\end{itemize}

\begin{figure*}[t]
    \centering
    \includegraphics[width=\linewidth]{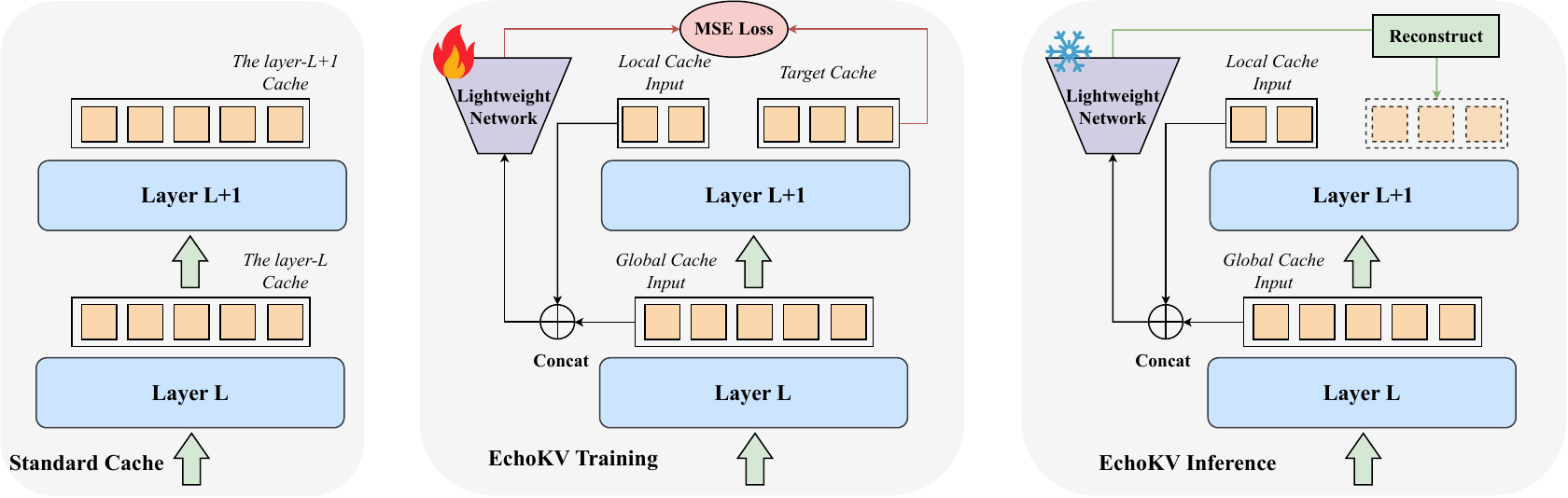}
    \caption{Schematic illustration of the training and inference workflows for EchoKV compared to the standard KV cache.
The figure presents a schematic illustration for a single token, where distinct cache blocks correspond to different segments of the flattened KV representation.
}
    \label{fig:main}
\end{figure*}

\section{Related work}

\subsection{KV cache compression}

\paragraph{Quantization.}
KV cache quantization methods are akin to existing activation quantization techniques, yet they exhibit unique characteristics \citep{dong2024qaq}.
Based on observations of KV cache distributions, KIVI \citep{liu2024kivi} adopts a combination of per-channel quantization for Keys and per-token quantization for Values,
achieving 2-bit lossless performance.
KVQuant \citep{hooper2024kvquant} defines quantization levels using a non-uniform distribution learned from calibration data, further enhancing information density.
AQUA-KV \citep{shutova2025cache} uses lightweight predictors to exploit KV dependencies and stores quantized prediction residuals.
Its predictors use the previous reconstructed layer, with Value prediction additionally using the current reconstructed Keys.
EchoKV instead reconstructs discarded dimensions from a full group-anchor cache and retained target-layer dimensions, without storing residuals for the discarded dimensions.
It is worth noting that quantization methods are orthogonal to EchoKV.
By further quantizing the retained KV cache to low-bit precision, 
we can achieve even higher compression ratios.

\paragraph{Dimensionality reduction.}
Beyond compressing data bit-width,
recent work \citep{shi2024keep} has also focused on directly compressing the dimensions of the KV cache.
Architecturally, Multi-Query Attention (MQA) \citep{shazeer2019fast} and Grouped-Query Attention (GQA) \citep{ainslie2023gqa} serve as explicit dimensional compression mechanisms for the KV cache.
Furthermore, Multi-Head Latent Attention (MLA) \citep{liu2024deepseek} replaces explicit grouping with implicit grouping,
utilizing low-rank projections to achieve significantly lower KV cache memory usage.
Recent efforts further convert existing pre-trained attention modules into MLA-style architectures, reducing KV cache memory but still requiring substantial continued training to recover model performance \citep{meng2025transmla,li2025xecomla}.

In addition to architectural improvements that require pre-training from scratch, 
recent work has also explored lightweight methods to compress KV cache dimensions.
Palu \citep{chang2024palu} employs inter-group SVD decomposition and matrix fusion techniques, compressing the cache footprint while maintaining inference efficiency.
CommonKV \citep{wang2025commonkv} extends this approach by applying SVD for inter-layer sharing, yielding better results even with limited memory budgets.

\subsection{KV cache eviction}
Based on the observation \citep{liu2023scissorhands} that a small subset of important tokens consistently receives subsequent attention, 
KV cache eviction methods \citep{zhang2023h2o} significantly reduce cache occupancy in long-context scenarios.
SnapKV \citep{li2024snapkv} evaluates token importance by using the last block of the input as an observation window.
LAQ \citep{wang2025lookahead} achieves importance assessment aligned with the generation phase by rapidly generating a pseudo-response as the observation window.
Judge-Q \citep{liu2025judge} further trains a set of soft tokens to serve as an observation window, achieving more precise importance assessment.
Since eviction methods operate at the granularity of individual tokens, they are orthogonal to EchoKV and can be combined with our method.

\section{Methodology}

\subsection{Overview of EchoKV}
As illustrated in Figure \ref{fig:main}, the core idea of EchoKV is to retain a partial KV representation and use a lightweight network to reconstruct the discarded components on demand.
By bypassing the explicit compression phase, our approach allows for directly evicting a portion of the KV cache when memory bottlenecks are encountered, and subsequently reconstructing it during inference using the retained KV information.
Serving as the core component of the reconstruction process, the lightweight network is detailed in this section, covering both its model architecture (\S\ref{sec:arc}) and training specifications (\S\ref{sec:train}).

\subsection{Network architecture}
\label{sec:arc}
Motivated by the effectiveness of SVD-based projections in prior KV compression methods,
we adopt \textbf{a simple linear layer} $\mathbf{W}_i$ as the default prediction network.
This choice prioritizes inference efficiency by keeping the architecture lightweight.
Leveraging the inter-layer and intra-layer similarity of the KV cache,
the network input is composed of two parts:
the full flattened KV representation of a specific layer serving as the global input,
and a retained partial representation of the current layer acting as the local input.
Aligning with prior works \citep{chang2024palu,wang2025commonkv},
we utilize keys prior to Rotary Positional Embedding (RoPE) \citep{su2024roformer} as inputs to enhance model adaptation.
Keys and Values use separate predictors with the same structure; $\mathbf{C}$ denotes either cache.
For each token, we flatten the head and feature dimensions of the KV cache into a single feature axis and perform reconstruction on this flattened representation.
Predictor dimensions are detailed in Appendix \ref{app:hyper_config}, Table \ref{tab:config_summary}.

\paragraph{Global cache inputs.}
To leverage inter-layer similarity, we partition the KV cache layers into distinct groups.
Within each group, the first layer retains its full cache to serve as global information, facilitating the prediction of the remaining layers in that group.
Formally, the global input for the $k$-th group is defined as:
\begin{equation}
    \mathbf{C}_{\text{global}}^{(k)} = \mathbf{C}_{k \cdot S},
    \label{eq:global}
\end{equation}
where $S$ represents the group size (i.e., the number of layers in each group), and $\mathbf{C}_{k \cdot S}$ is the full KV cache of the first layer in that group.

\paragraph{Local cache inputs.}
Furthermore, to mitigate potential information loss across layers,
we incorporate a retained subset of flattened KV dimensions from the current layer as the local input,
thereby enhancing prediction accuracy.
For the $i$-th layer within the $k$-th group, the local cache input can be formulated as:
\begin{equation}
    \mathbf{C}_{\text{local}}^{(k,i)} = \mathbf{C}_{k \cdot S+i}\text{[:, :m]},
\end{equation}
where $\mathbf{C}_{k \cdot S+i}$ is the full cache of the layer,
and $m$ is the number of retained local feature dimensions used to preserve intra-layer details.
Considering the efficiency of contiguous memory access,
we directly retain the first $m$ flattened dimensions as the local input based on the target budget, without employing complex selection heuristics.
Appendix \ref{app:sm_ablation} and Appendix \ref{app:local_select} further examine the effects of the group size $S$, the retained local dimension $m$, and alternative local-dimension selection strategies.

Synthesizing the above,
we concatenate the global and local features of each token and employ the lightweight model $\mathbf{W}_{k \cdot S + i}$ to predict the evicted cache.
This process can be formulated as:
\begin{equation}
    \hat{\mathbf{C}}^{(k,i)}_{\text{drop}} = \mathbf{W}_{k \cdot S + i} \left[ \mathbf{C}_{\text{global}}^{(k)} \ ; \ \mathbf{C}_{\text{local}}^{(k,i)} \right].
\end{equation}
The ground truth for the prediction corresponds to the discarded flattened dimensions $\mathbf{C}^{(k,i)}_{\text{drop}}$ that are evicted during inference:
\begin{equation}
    \mathbf{C}^{(k,i)}_{\text{drop}} = \mathbf{C}_{k \cdot S+i}\text{[:, m:]}.
\end{equation}
When $m=0$, the group's first layer supplies the entire predictor input.
We choose the smallest group whose first-layer cache fits the target budget, then allocate the remaining budget to local dimensions.
Appendix \ref{app:sm_ablation} gives the budget equations and motivation; Appendix \ref{app:config} lists the configurations.

\paragraph{Inference and switching.}
Prefill attention uses full K/V while storing the compact cache; decoding reconstructs dropped dimensions before attention.
An existing FullKV cache can be compacted mid-session without recomputing the prefix.
Appendix \ref{app:inference} describes the procedure and switching protocol, with results in Appendix \ref{app:switching_results}.

\subsection{Training details}
\label{sec:train}
To fully train the lightweight network and ensure the usability of the reconstructed cache,
we employ a two-stage training strategy to progressively enhance the performance of EchoKV.
\paragraph{Reconstruction loss.}
In the initialization stage, we directly employ the reconstruction MSE loss computed on the corresponding Key-Value pairs for training.
This can be formulated as:
\begin{equation}
    \mathcal{L}_{\text{init}} = \left\| \hat{\mathbf{C}}_{\text{drop}} - \mathbf{C}_{\text{drop}} \right\|_2^2.
\end{equation}
In this phase, we strive for a faithful reconstruction of the KV cache, which serves as a robust weight initialization for the subsequent training stage.
However, we observe that relying solely on reconstruction yields suboptimal performance in long-context scenarios (as detailed in \S \ref{sec:loss}).
We attribute this limitation to the fact that pure reconstruction loss does not directly optimize the downstream attention output.
Consequently, we proceed to a second training stage to further refine the network under actual attention mechanisms.

\paragraph{Attention loss.}
Formally, the reconstructed Key and Value representations for the $i$-th layer within the $k$-th group are obtained by concatenating the retained local features and the predicted features, followed by reshaping them back to the original KV layout:
\begin{equation}
\label{eq:concat}
\tilde{\mathbf{X}}_{k,i} = \mathrm{Reshape}\left(\left[ \mathbf{X}^{(k,i)}_{\text{local}} \ ; \ \hat{\mathbf{X}}^{(k,i)}_{\text{drop}} \right]\right), \quad \mathbf{X} \in \{\mathbf{K}, \mathbf{V}\},
\end{equation}
where $[\cdot \ ; \ \cdot]$ denotes the concatenation operation,
$\mathbf{X}^{(k,i)}_{\text{local}}$ denotes the retained local features, and $\hat{\mathbf{X}}^{(k,i)}_{\text{drop}}$ denotes the predicted components for $\mathbf{X} \in \{\mathbf{K}, \mathbf{V}\}$.

To ensure computational efficiency in long-context scenarios, we eschew the standard KL divergence loss, which typically requires materializing the full attention matrix to align $\mathbf{Q}\mathbf{K}^T$ and $\mathbf{Q}\tilde{\mathbf{K}}^T$
(detailed in Appendix \ref{app:qk_loss}).
Instead, we adopt an Output MSE (O-MSE) loss that is fully compatible with FlashAttention \citep{dao2022flashattention}.
This approach facilitates efficient training while jointly optimizing both the reconstructed Keys ($\tilde{\mathbf{K}}$) and Values ($\tilde{\mathbf{V}}$), as formulated below:
\begin{equation}
    \label{eq:o_loss}
    \mathcal{L}_{\text{attention}} = \left\| \mathcal{A}(\mathbf{Q}, \mathbf{K}, \mathbf{V}) - \mathcal{A}(\mathbf{Q}, \tilde{\mathbf{K}}, \tilde{\mathbf{V}}) \right\|_2^2,
\end{equation}
where $\mathcal{A}(\cdot)$ denotes the FlashAttention operation.

It is worth noting that, despite the adoption of a two-stage training strategy, the training process remains highly efficient and can be completed within a very short timeframe.
This is attributed to the lightweight nature of our network,
which contains up to approximately 50M parameters for Llama-3.1-8B (detailed in Table \ref{tab:config_summary}).
Specifically, we can complete the training of EchoKV for Llama-3.1-8B in only a few minutes on a single A100 GPU.

\begin{table*}[t]
\fontsize{22}{26}\selectfont
\centering
\caption{Experimental results of different models and methods on LongBench.
$^*$ Since ThinK only compresses Keys, we double the compression rate calculation to account for Value storage. We report results at 0.6 because the 0.5 ratio is infeasible.
Compression Ratio = Size of Compressed Cache / Size of Full Cache.
}
\label{tab:longbench}
\begin{threeparttable}
\setlength{\tabcolsep}{8pt}
\scalebox{0.285}{
\begin{tabular}{llccccccccccccccccc}
\toprule[2pt]
&\multirow{4}{*}{\textbf{~~~Methods}} & \multicolumn{3}{c}{\textbf{Single-Document QA}} & \multicolumn{3}{c}{\textbf{Multi-Document QA}}& \multicolumn{3}{c}{\textbf{Summarization}}& \multicolumn{3}{c}{\textbf{Few-shot Learning}}& \multicolumn{2}{c}{\textbf{Synthetic}} & \multicolumn{2}{c}{\textbf{Code}}&\multirow{4}{*}{\textbf{~~~Avg.}} \\
\cmidrule(lr){3-5}\cmidrule(lr){6-8}\cmidrule(lr){9-11}\cmidrule(lr){12-14}\cmidrule(lr){15-16}\cmidrule(lr){17-18}
&& \rotatebox[origin=c]{30}{\textbf{Nrtv}} & \rotatebox[origin=c]{30}{\textbf{Qasp}} & \rotatebox[origin=c]{30}{\textbf{MF}} & \rotatebox[origin=c]{30}{\textbf{Hotpot}} & \rotatebox[origin=c]{30}{\textbf{2Wiki}} & \rotatebox[origin=c]{30}{\textbf{Musiq}} & \rotatebox[origin=c]{30}{\textbf{Gov}} & \rotatebox[origin=c]{30}{\textbf{QMS}} & \rotatebox[origin=c]{30}{\textbf{MN}} & \rotatebox[origin=c]{30}{\textbf{TREC}} & \rotatebox[origin=c]{30}{\textbf{TQA}} & \rotatebox[origin=c]{30}{\textbf{SAM}} & \rotatebox[origin=c]{30}{\textbf{PC}} & \rotatebox[origin=c]{30}{\textbf{PR}} & \rotatebox[origin=c]{30}{\textbf{LCC}} & \rotatebox[origin=c]{30}{\textbf{RB}} \\

\midrule
\multirow{16}{*}{\rotatebox[origin=c]{90}{\fontsize{18}{100}\selectfont \textbf{Llama-3.1-8B-Instruct}}}

&~~~FullKV & 31.38 & 46.60 & 56.85 & 58.10 & 50.01 & 32.52 & 34.66 & 25.19 & 27.11 & 73.00 & 92.11 & 43.70 & 6.56 & 100.0 & 65.52 & 54.75 & 49.88 \\
\cline{2-19}

& \multicolumn{18}{c}{\cellcolor{lightgray!25} 
  \textit{Compression Ratio = 0.7}
} \\

&~~~MiniCache & 17.87 & 22.26 & 29.51 & 26.72 & 11.47 & 9.71 & 24.61 & 21.85 & 23.29 & 65.00 & 75.05 & 31.46 & 5.83 & 60.50 & 41.34 & 37.19 & 31.48 \\

&~~~ThinK & \textbf{32.77} & 45.44 & \textbf{58.83} & \textbf{58.30} & \textbf{51.09} & \textbf{32.65} & \textbf{32.63} & 24.96 & \textbf{26.19} & \textbf{73.00} & 91.72 & 42.10 & \textbf{6.96} & 99.50 & \textbf{65.94} & 55.64 & \textbf{49.86} \\

&~~~Palu & 28.96 & \textbf{46.73} & 53.26 & 53.08 & 45.35 & 29.91 & 32.51 & 24.69 & 25.42 & \textbf{73.00} & 89.10 & \textbf{43.77} & 2.50 & 96.50 & 57.90 & \textbf{56.78} & 47.47 \\

&~~~CommonKV & 29.87 & 41.16 & 53.86 & 54.90 & 45.42 & 29.96 & 27.91 & 24.21 & 25.68 & 66.00 & 90.03 & 43.10 & 6.67 & \textbf{100.0} & 61.92 & 53.41 & 47.13 \\

&~~~EchoKV & 32.24 & 44.53 & 57.84 & 57.90 & 49.89 & 32.08 & 29.29 & \textbf{25.06} & 25.02 & \textbf{73.00} & \textbf{92.52} & 41.67 & 6.87 & 99.00 & 63.57 & 54.72 & 49.08 \\

\cline{2-19}
& \multicolumn{18}{c}{\cellcolor{lightgray!25} 
  \textit{Compression Ratio = 0.5}
} \\

&~~~ThinK* & 28.51 & 22.92 & 37.42 & 52.90 & 47.84 & 28.01 & 19.67 & 21.96 & 19.01 & 38.00 & 88.11 & 40.72 & 7.50 & \textbf{99.50} & 59.97 & 49.92 & 41.37 \\

&~~~Palu & 19.27 & 34.64 & 40.14 & 41.99 & 26.71 & 22.45 & 22.60 & 23.70 & 23.21 & 66.00 & 25.95 & 39.75 & 4.00 & 49.50 & 29.86 & 30.19 & 31.25 \\

&~~~CommonKV & 25.24 & 40.64 & 54.45 & 54.56 & 45.58 & 29.82 & 25.23 & 23.85 & 24.92 & 59.00 & 89.98 & \textbf{42.50} & \textbf{10.12} & \textbf{99.50} & 62.46 & 52.50 & 46.27 \\

&~~~EchoKV & \textbf{33.06} & \textbf{44.80} & \textbf{57.51} & \textbf{57.91} & \textbf{50.25} & \textbf{31.79} & \textbf{29.22} & \textbf{24.10} & \textbf{25.21} & \textbf{73.00} & \textbf{91.92} & 41.45 & 7.17 & 89.50 & \textbf{64.61} & \textbf{54.91} & \textbf{48.53} \\

\cline{2-19}
& \multicolumn{18}{c}{\cellcolor{lightgray!25} 
  \textit{Compression Ratio = 0.3}
} \\
&~~~Palu & 1.90 & 2.42 & 4.36 & 1.00 & 1.70 & 0.66 & 3.52 & 9.28 & 5.95 & 1.25 & 3.19 & 6.31 & 0.00 & 0.00 & 18.20 & 18.79 & 4.91 \\

&~~~CommonKV & 11.20 & 29.58 & 36.89 & 31.21 & 31.30 & 14.79 & 13.86 & 21.22 & 21.82 & 47.50 & 83.38 & 40.25 & 0.22 & 8.50 & 54.43 & 51.10 & 31.08 \\

&~~~EchoKV & \textbf{31.22} & \textbf{42.47} & \textbf{54.27} & \textbf{57.03} & \textbf{49.58} & \textbf{30.75} & \textbf{24.23} & \textbf{23.84} & \textbf{22.32} & \textbf{49.00} & \textbf{91.79} & \textbf{40.90} & \textbf{7.25} & \textbf{84.50} & \textbf{62.09} & \textbf{53.03} & \textbf{45.27} \\

\midrule

\multirow{16}{*}{\rotatebox[origin=c]{90}{\fontsize{18}{100}\selectfont \textbf{Mistral-7B-Instruct-v0.3}}}

&~~~FullKV & 29.80 & 39.12 & 50.45 & 50.28 & 36.31 & 26.45 & 34.11 & 25.93 & 26.56 & 76.00 & 88.89 & 46.97 & 5.50 & 97.00 & 61.47 & 62.57 & 47.34 \\
\cline{2-19}

& \multicolumn{18}{c}{\cellcolor{lightgray!25} 
  \textit{Compression Ratio = 0.7}
} \\

&~~~MiniCache & 14.49 & 21.16 & 26.26 & 18.38 & 18.16 & 6.33 & 19.12 & 21.06 & 23.22 & 61.00 & 82.68 & 32.60 & 2.67 & 24.17 & 44.28 & 40.75 & 28.52 \\

&~~~ThinK & \textbf{30.42} & \textbf{37.87} & 50.13 & 50.35 & 34.55 & 25.92 & \textbf{33.81} & \textbf{25.66} & 26.40 & \textbf{76.00} & 88.43 & 45.81 & \textbf{5.50} & 95.00 & 61.90 & 61.95 & 46.86 \\

&~~~Palu & 28.47 & 36.08 & \textbf{52.56} & 46.57 & \textbf{36.12} & \textbf{26.37} & 33.59 & 25.16 & \textbf{26.54} & 73.50 & 87.77 & 45.11 & 4.00 & 94.50 & 59.04 & 61.06 & 46.03 \\

&~~~CommonKV & 26.02 & 37.13 & 48.21 & 47.61 & 29.87 & 24.43 & 30.05 & 24.75 & 25.76 & 74.00 & 88.12 & 44.54 & 4.00 & 92.50 & 61.16 & 62.10 & 45.02 \\

&~~~EchoKV & 29.93 & 37.65 & 49.43 & \textbf{51.50} & 34.54 & 24.98 & 32.29 & 25.11 & 25.99 & \textbf{76.00} & \textbf{89.47} & \textbf{47.44} & \textbf{5.50} & \textbf{95.50} & \textbf{62.26} & \textbf{62.18} & \textbf{46.86} \\

\cline{2-19}
& \multicolumn{18}{c}{\cellcolor{lightgray!25} 
  \textit{Compression Ratio = 0.5}
} \\

&~~~ThinK* & 29.27 & 27.37 & 39.90 & 45.95 & 31.83 & 20.41 & 22.24 & 23.21 & 19.80 & 65.50 & 87.27 & 43.43 & \textbf{6.00} & 80.00 & 59.03 & 59.32 & 41.28 \\

&~~~Palu & 25.84 & 35.28 & 47.43 & 46.69 & 31.93 & \textbf{27.21} & 29.32 & 24.32 & \textbf{25.37} & 74.50 & 86.31 & 42.95 & 4.50 & 61.00 & 48.19 & 47.90 & 41.17 \\

&~~~CommonKV & 27.27 & 34.71 & 46.70 & 44.74 & 27.02 & 24.21 & 26.36 & 23.52 & 24.42 & 51.00 & 88.56 & 43.82 & 4.50 & 88.00 & 60.14 & 59.94 & 42.18 \\

&~~~EchoKV & \textbf{30.26} & \textbf{36.92} & \textbf{49.71} & \textbf{51.33} & \textbf{34.26} & 24.51 & \textbf{31.54} & \textbf{25.28} & 25.36 & \textbf{76.00} & \textbf{89.32} & \textbf{46.93} & 3.50 & \textbf{93.50} & \textbf{61.72} & \textbf{61.66} & \textbf{46.36} \\

\cline{2-19}
& \multicolumn{18}{c}{\cellcolor{lightgray!25} 
  \textit{Compression Ratio = 0.3}
} \\

&~~~Palu & 11.04 & 11.91 & 23.13 & 16.00 & 14.48 & 8.77 & 10.60 & 20.50 & 16.87 & 58.50 & 58.98 & 26.55 & \textbf{4.50} & 4.08 & 23.34 & 25.56 & 20.93 \\

&~~~CommonKV & 11.96 & 21.25 & 29.05 & 21.47 & 14.49 & 8.40 & 22.30 & 22.55 & \textbf{24.43} & 56.50 & 72.81 & 37.21 & 2.63 & 14.50 & 52.46 & 39.16 & 28.20 \\

&~~~EchoKV & \textbf{27.99} & \textbf{35.01} & \textbf{47.29} & \textbf{49.77} & \textbf{31.63} & \textbf{24.01} & \textbf{25.88} & \textbf{23.86} & 23.63 & \textbf{67.00} & \textbf{89.06} & \textbf{45.00} & 4.00 & \textbf{70.00} & \textbf{59.27} & \textbf{61.05} & \textbf{42.78} \\

\midrule

\multirow{16}{*}{\rotatebox[origin=c]{90}{\fontsize{18}{100}\selectfont \textbf{Mistral-Small-24B-Instruct-2501}}}

&~~~FullKV & 33.72 & 47.19 & 54.95 & 67.24 & 64.93 & 49.04 & 33.22 & 24.54 & 25.47 & 75.50 & 93.71 & 48.86 & 21.00 & 100.0 & 63.67 & 70.66 & 54.61 \\
\cline{2-19}

& \multicolumn{18}{c}{\cellcolor{lightgray!25}
  \textit{Compression Ratio = 0.7}
} \\

&~~~MiniCache & 31.70 & 35.78 & 44.52 & 61.25 & 53.13 & 39.06 & 25.38 & 22.30 & 22.92 & 72.00 & 92.39 & 37.20 & 17.50 & 67.50 & 37.97 & 48.19 & 44.30 \\

&~~~ThinK & 34.43 & 41.31 & 51.85 & \textbf{67.44} & \textbf{64.09} & 47.23 & 24.61 & 23.98 & 20.62 & 61.00 & 93.24 & 46.86 & 19.00 & \textbf{100.0} & 61.30 & 69.05 & 51.63 \\

&~~~Palu & 34.71 & \textbf{47.09} & 55.35 & 66.61 & 63.78 & 44.51 & \textbf{32.68} & \textbf{25.72} & 25.53 & 74.50 & 92.12 & \textbf{48.22} & 14.50 & \textbf{100.0} & 54.65 & 65.63 & 52.85 \\

&~~~CommonKV & \textbf{36.16} & 46.59 & 54.82 & 66.00 & 62.80 & 46.29 & 32.13 & 24.85 & \textbf{25.55} & \textbf{76.00} & 93.16 & 48.20 & 18.00 & 99.50 & \textbf{68.02} & \textbf{71.11} & 54.32 \\

&~~~EchoKV & 33.74 & 46.73 & \textbf{55.47} & 67.18 & 64.07 & \textbf{49.50} & 31.71 & 25.28 & 24.92 & 75.50 & \textbf{93.64} & \textbf{48.22} & \textbf{19.50} & \textbf{100.0} & 64.38 & 70.98 & \textbf{54.43} \\

\cline{2-19}
& \multicolumn{18}{c}{\cellcolor{lightgray!25}
  \textit{Compression Ratio = 0.5}
} \\

&~~~ThinK* & 26.96 & 30.79 & 35.81 & 56.43 & 57.52 & 39.26 & 19.81 & 21.56 & 16.84 & 37.00 & 91.51 & 44.48 & 17.50 & 99.00 & 56.30 & 61.10 & 44.49 \\

&~~~Palu & 28.82 & 44.48 & 47.35 & 61.33 & 53.18 & 37.76 & 23.24 & \textbf{24.79} & 22.87 & 73.00 & 91.12 & 43.68 & 9.50 & 97.00 & 22.62 & 38.07 & 44.93 \\

&~~~CommonKV & \textbf{35.98} & 46.14 & 54.49 & 66.59 & 63.45 & 47.10 & 30.78 & 24.58 & \textbf{25.52} & 74.00 & 92.91 & 47.59 & 19.00 & \textbf{100.0} & \textbf{66.90} & \textbf{70.02} & 54.07 \\

&~~~EchoKV & 32.78 & \textbf{46.33} & \textbf{56.23} & \textbf{67.14} & \textbf{64.25} & \textbf{49.07} & \textbf{30.88} & 24.69 & 24.38 & \textbf{75.50} & \textbf{93.39} & \textbf{48.37} & \textbf{20.00} & \textbf{100.0} & 64.14 & 69.87 & \textbf{54.19} \\

\cline{2-19}
& \multicolumn{18}{c}{\cellcolor{lightgray!25}
  \textit{Compression Ratio = 0.3}
} \\

&~~~Palu & 4.77 & 6.85 & 16.03 & 7.47 & 5.98 & 6.47 & 7.06 & 15.09 & 10.45 & 36.50 & 22.94 & 16.84 & 1.00 & 0.00 & 21.35 & 22.41 & 12.58 \\

&~~~CommonKV & 27.41 & 41.87 & 48.97 & 55.45 & 53.02 & 28.62 & 26.16 & 22.74 & \textbf{24.72} & 72.50 & 90.17 & 45.51 & 1.23 & 95.00 & \textbf{70.49} & 68.78 & 48.29 \\

&~~~EchoKV & \textbf{30.77} & \textbf{44.80} & \textbf{54.43} & \textbf{66.78} & \textbf{62.74} & \textbf{49.01} & \textbf{27.57} & \textbf{24.66} & 23.05 & \textbf{74.50} & \textbf{93.23} & \textbf{45.76} & \textbf{18.50} & \textbf{99.50} & 63.25 & \textbf{69.95} & \textbf{53.03} \\

\bottomrule[2pt]
\end{tabular}
}
\end{threeparttable}
\vspace{-10pt}
\end{table*}

\section{Experiments}

\subsection{Experimental setting}

\paragraph{Datasets.}
To comprehensively evaluate the effectiveness of our compression method,
we conduct experiments on both the real-world long-context benchmark LongBench \citep{bai2024longbench}
and the synthetic dataset RULER \citep{hsieh2024ruler}.
Specifically, we utilize the maximum context length of models for LongBench,
while using 32K sequences for the main RULER evaluation.
Additional 64K/128K tests and predictor-seed evaluations are reported in Appendices \ref{app:ultralong} and \ref{app:seed_robustness}.

Regarding the training dataset, we explore a diverse range of options.
As the results in \S \ref{sec:dataset} indicate,
the proposed method is insensitive to the specific choice of training data.
We ultimately employ LongAlpaca \citep{long-alpaca} for our experiments.

\paragraph{Baselines.}
We select \textit{Palu} \citep{chang2024palu} and \textit{CommonKV} \citep{wang2025commonkv}, two SVD-based state-of-the-art (SOTA) methods, as strong baselines.
Furthermore, we compare our approach with the low-rank method \textit{ThinK} \citep{xu2024think} and the post-compression merging technique \textit{MiniCache} \citep{liu2024minicache},
and validate the potential for integrating these methods with our proposed \textit{EchoKV}.
We also include \textit{TransMLA} \citep{meng2025transmla}, which converts existing attention modules into an MLA-style architecture, as an additional architecture-conversion baseline.
For the backbone models, we employ Llama-3.1-8B-Instruct \citep{grattafiori2024llama}, Mistral-7B-Instruct-v0.3 \citep{Jiang2023Mistral7}, and Mistral-Small-24B-Instruct-2501,
covering model scales from 7B to 24B.
\paragraph{Implementation details.}
To ensure the transferability of our method,
we avoid specific hyperparameter search or tuning for layer-wise budgets.
Instead, we use the same group size $S$ and retained local dimension $m$ across layer groups
(detailed in Appendix \ref{app:config}).
Considering the impact of attention sinks \citep{xiao2023efficient},
we follow established baselines by retaining 4 sink tokens and a local window of 128 tokens at full precision.
Reported target ratios describe the compressed token region; dimensional rounding and full-token storage are detailed in Appendix \ref{app:config}.
Appendix \ref{app:hyperpara} presents the hyperparameters for the lightweight network training stage.

We also evaluate an exploratory hybrid variant that combines low-rank compression for Keys with EchoKV reconstruction for Values, motivated by their asymmetric reconstruction difficulty.
Detailed hybrid results and discussion are provided in Appendix \ref{app:hybrid_results}.
We further defer several implementation ablations to Appendix \ref{app:sm_ablation}, Appendix \ref{app:local_select}, and Appendix \ref{app:pred_arch}, covering the choices of $(S, m)$, local-dimension selection strategy, and predictor architecture.

\begin{table*}[t]

\fontsize{20}{26}\selectfont
\setlength{\tabcolsep}{12pt}
\centering
\caption{
Experimental results of different methods on RULER at a compression ratio of 0.5.
$^*$ Indicates the actual compression ratio is 0.6, consistent with Table \ref{tab:longbench}.
}\label{tab:ruler_0.5}
\begin{threeparttable}
\scalebox{0.345}{
\begin{tabular}{lcccccccccccccc}
\toprule[2pt]
\multirow{4}{*}{\textbf{Method}} & \multicolumn{8}{c}{\textbf{Needle-in-a-Haystack (NIAH)}} & \multicolumn{3}{c}{\textbf{Synthetic}}
& \multicolumn{2}{c}{\textbf{QA}}
&\multirow{4}{*}{\textbf{~~~Avg.}} \\
\cmidrule(lr){2-9}\cmidrule(lr){10-12}\cmidrule(lr){13-14}
& \rotatebox[origin=c]{30}{\textbf{S1}} & \rotatebox[origin=c]{30}{\textbf{S2}} & \rotatebox[origin=c]{30}{\textbf{S3}} & \rotatebox[origin=c]{30}{\textbf{MK1}} & \rotatebox[origin=c]{30}{\textbf{MK2}} & \rotatebox[origin=c]{30}{\textbf{MK3}} & \rotatebox[origin=c]{30}{\textbf{MV}} & \rotatebox[origin=c]{30}{\textbf{MQ}} & \rotatebox[origin=c]{30}{\textbf{VT}} & \rotatebox[origin=c]{30}{\textbf{CWE}} & \rotatebox[origin=c]{30}{\textbf{FWE}} & \rotatebox[origin=c]{30}{\textbf{QA1}} & \rotatebox[origin=c]{30}{\textbf{QA2}} & \\
\midrule

\multicolumn{15}{c}{\cellcolor{lightgray!25} \textit{Llama-3.1-8B-Instruct}} \\
FullKV & 100.0 & 100.0 & 100.0 & 99.00 & 99.00 & 99.00 & 98.75 & 98.50 & 99.20 & 17.80 & 87.00 & 87.00 & 54.00 & 87.63 \\
\hdashline
ThinK* & 1.00 & 2.00 & 0.00 & 2.00 & 0.00 & 0.00 & 1.75 & 3.25 & 0.60 & 0.00 & 48.00 & 63.00 & 40.00 & 12.43 \\
Palu & \textbf{100.0} & 98.00 & 75.00 & 97.00 & 78.00 & 30.00 & 73.00 & 64.25 & 86.00 & \textbf{15.30} & 78.33 & 53.00 & 38.00 & 68.14 \\
CommonKV & 92.00 & 94.00 & 83.00 & 97.00 & 67.00 & 1.00 & 87.00 & 94.25 & 79.80 & 0.10 & \textbf{85.67} & 75.00 & 50.00 & 69.68 \\
EchoKV & \textbf{100.0} & \textbf{100.0} & \textbf{100.0} & \textbf{99.00} & \textbf{99.00} & \textbf{94.00} & \textbf{89.50} & \textbf{99.50} & \textbf{87.20} & 0.60 & 80.00 & \textbf{85.00} & \textbf{52.00} & \textbf{83.52} \\

\midrule
\multicolumn{15}{c}{\cellcolor{lightgray!25} 
  \textit{Mistral-7B-Instruct-v0.3}
} \\
FullKV & 100.0 & 92.00 & 100.0 & 84.00 & 80.00 & 57.00 & 92.25 & 92.25 & 95.80 & 80.70 & 89.67 & 65.00 & 47.00 & 82.74 \\
\hdashline
ThinK* & 21.00 & 58.00 & 2.00 & 32.00 & 17.00 & 5.00 & 20.25 & 14.00 & 44.20 & 17.10 & \textbf{89.67} & 60.00 & 42.00 & 32.47 \\
Palu & \textbf{100.0} & \textbf{97.00} & 98.00 & 76.00 & 51.00 & \textbf{26.00} & \textbf{88.50} & \textbf{80.75} & 83.00 & 32.10 & 85.00 & 41.00 & 33.00 & 68.56 \\
CommonKV & 86.00 & 46.00 & 36.00 & 27.00 & 18.00 & 0.00 & 7.00 & 12.75 & 66.60 & 15.10 & 75.00 & 46.00 & \textbf{44.00} & 36.88 \\
EchoKV & \textbf{100.0} & 88.00 & \textbf{99.00} & \textbf{82.00} & \textbf{59.00} & 19.00 & 62.75 & 67.50 & \textbf{85.20} & \textbf{46.60} & 81.67 & \textbf{65.00} & 43.00 & \textbf{69.13} \\
\midrule
\multicolumn{15}{c}{\cellcolor{lightgray!25}
  \textit{Mistral-Small-24B-Instruct-2501}
} \\
FullKV & 98.00 & 100.0 & 100.0 & 95.00 & 98.00 & 94.00 & 99.75 & 99.00 & 100.0 & 95.90 & 92.00 & 83.00 & 63.00 & 93.67 \\
\hdashline
ThinK* & 0.00 & 1.00 & 1.00 & 1.00 & 0.00 & 0.00 & 2.75 & 2.50 & 0.40 & 12.80 & 25.33 & 59.00 & 46.00 & 11.68 \\
Palu & 98.00 & \textbf{100.0} & 68.00 & 97.00 & 84.00 & 76.00 & \textbf{94.50} & 96.25 & 71.20 & 34.20 & 80.00 & 55.00 & 58.00 & 77.86 \\
CommonKV & \textbf{99.00} & 99.00 & 98.00 & \textbf{98.00} & 92.00 & 90.00 & \textbf{94.50} & \textbf{98.00} & 89.60 & 71.10 & \textbf{94.00} & 81.00 & \textbf{62.00} & 89.71 \\
EchoKV & \textbf{99.00} & \textbf{100.0} & \textbf{100.0} & 95.00 & \textbf{98.00} & \textbf{95.00} & 85.00 & 96.25 & \textbf{98.00} & \textbf{75.20} & 92.33 & \textbf{83.00} & 60.00 & \textbf{90.52} \\
\bottomrule[2pt]
\end{tabular}
}
\end{threeparttable}\vspace{-10pt}
\end{table*}

\subsection{Main results}

\paragraph{Results on LongBench.}
We evaluate the performance of various baseline methods across multiple models on LongBench \citep{bai2024longbench},
as shown in Table \ref{tab:longbench}.
By capturing latent similarities in KV representations, EchoKV achieves competitive performance, with advantages at lower cache budgets.
At ratio 0.7, ThinK scores higher on Llama-3.1-8B (49.86 vs.\ 49.08) and ties EchoKV on Mistral-7B (46.86).
While methods based on parameter SVD decomposition mitigate performance degradation, they fail to enable flexible switching during inference.
Using both local and global KV information, EchoKV retains LongBench averages close to FullKV at ratio 0.5 and outperforms the evaluated compression baselines at 0.3.
These results support a setting-dependent quality--cache trade-off alongside flexible inference switching.
Notably, KV compression is typically orthogonal to eviction and quantization \citep{chang2024palu,wang2025commonkv}, 
allowing these techniques to be combined to achieve even higher compression ratios.
We also report an exploratory hybrid variant in Appendix \ref{app:hybrid_results}, where heterogeneous treatment of Keys and Values yields further gains in several moderate-budget settings.

\begin{wraptable}{r}{0.52\linewidth}
    \vspace{-0.8em}
    \centering
    \scriptsize
    \setlength{\tabcolsep}{3.5pt}
    \caption{LongBench average scores compared with TransMLA. ``Converted'' denotes direct architecture conversion, while ``FT'' denotes additional training on the same 1.6K LongAlpaca samples used by EchoKV.}
    \label{tab:transmla}
    \begin{tabular}{lcccc}
    \toprule
    \textbf{Model} & \textbf{Ratio} & \textbf{Converted} & \textbf{FT} & \textbf{EchoKV} \\
    \midrule
    Llama-3.1-8B & 0.5 & 27.65 & 36.89 & \textbf{48.53} \\
    Llama-3.1-8B & 0.3 & 11.76 & 32.31 & \textbf{45.27} \\
    Mistral-7B & 0.5 & 21.15 & 29.88 & \textbf{46.36} \\
    Mistral-7B & 0.3 & 12.36 & 25.00 & \textbf{42.78} \\
    \bottomrule
    \end{tabular}
    \vspace{-0.8em}
\end{wraptable}

\paragraph{Results on RULER.}
Beyond the real-world datasets of LongBench, we also perform a detailed assessment on the synthetic long-context benchmark, RULER \citep{hsieh2024ruler}.
Table \ref{tab:ruler_0.5} presents the results at a compression ratio of 0.5, while additional results, including the hybrid variant, are reported in Appendix \ref{app:more_res}.
It can be observed that compared to robust real-world datasets, the differences between methods are more significant on synthetic tasks.
At ratio 0.5, EchoKV exceeds the evaluated compression baselines on average, with gaps to FullKV of 4.11, 13.61, and 3.15 points for Llama-3.1-8B, Mistral-7B, and Mistral-Small-24B, respectively.
The appendix further shows that the hybrid variant can bring additional gains on some RULER settings, although this advantage is not uniform across models and compression regimes.

At 64K/128K, EchoKV scores 79.63/62.54 on Llama-3.1-8B, compared with FullKV's 84.13/75.46 (Appendix \ref{app:ultralong}).
Across four predictor-training seeds on this model at ratios 0.7 and 0.5, standard deviations of benchmark averages are at most 0.14 on LongBench and 0.27 on RULER (Appendix \ref{app:seed_robustness}).

\paragraph{Comparison with TransMLA.}
Table \ref{tab:transmla} compares EchoKV with TransMLA \citep{meng2025transmla} on LongBench.
Direct architecture conversion causes severe degradation, and TransMLA requires substantial continued training to recover performance.
Even after additional training on the same 1.6K LongAlpaca samples used by EchoKV, it still remains clearly behind EchoKV across both backbones and compression ratios.
Detailed conversion settings and additional analysis are provided in Appendix \ref{app:transmla_results}.

\section{Analysis}

\subsection{Selection of objective functions}
\label{sec:loss}
As discussed in \S \ref{sec:train},
we employ a two-stage training strategy with distinct loss functions to optimize the lightweight network.
In this section, we analyze different loss functions to validate the effectiveness of our proposed method.

As shown in Table \ref{tab:loss}, optimizing O-MSE from scratch is less effective than first fitting the dropped KV entries with KV-MSE.
This suggests that the first stage mainly serves as a representation-alignment step: before the predictor learns a faithful cache reconstruction, the attention-space objective is too indirect to provide a strong optimization signal.
Once such an initialization is available, however, O-MSE becomes a better second-stage objective because it directly optimizes the downstream attention outputs.

\begin{wraptable}{r}{0.37\linewidth}
    \vspace{-1.2em}
    \centering
    \scriptsize
    \setlength{\tabcolsep}{3.5pt}
    \caption{Training time and performance across different loss functions.}
    \label{tab:loss}
    \begin{tabular}{cccc}
    \toprule
    \textbf{Loss} & \textbf{Stage} & \shortstack{\textbf{Total}\\\textbf{time}} & \textbf{LB Avg.}\\
    \midrule
    KV-MSE & \uppercase\expandafter{\romannumeral1} & 12.5 min & 48.99 \\
    O-MSE & \uppercase\expandafter{\romannumeral1} & 15.4 min & 48.76 \\
    \midrule
    QK-KL & \uppercase\expandafter{\romannumeral2} & 39.7 min & 49.11 \\
    O-MSE & \uppercase\expandafter{\romannumeral2} & 14.5 min & \textbf{49.26} \\
    \bottomrule
    \end{tabular}
    \vspace{-1.2em}
\end{wraptable}

We also compare O-MSE with the commonly used KL divergence loss on QK attention distributions (detailed in Appendix \ref{app:qk_loss}).
While QK-KL and O-MSE reach similar final performance, O-MSE is considerably more practical for long-context training.
Unlike QK-KL, it remains compatible with FlashAttention \citep{dao2022flashattention,dao2023flashattention2} and avoids explicit materialization of the $L \times L$ attention matrix.
In our setting, this reduces training time by nearly $3\times$ without sacrificing final quality, which makes the two-stage training procedure efficient in practice.
Appendix \ref{app:hyperpara} reports the complete training schedules and corresponding RULER results.

\begin{figure*}[t]
   \centering
   \subcaptionbox{Difficulty analysis of the EchoKV reconstruction task for keys and values.
   ``Save K'' denotes only values are reconstructed.
   \label{fig:analy:kv}}
   {\includegraphics[width=0.31\textwidth]{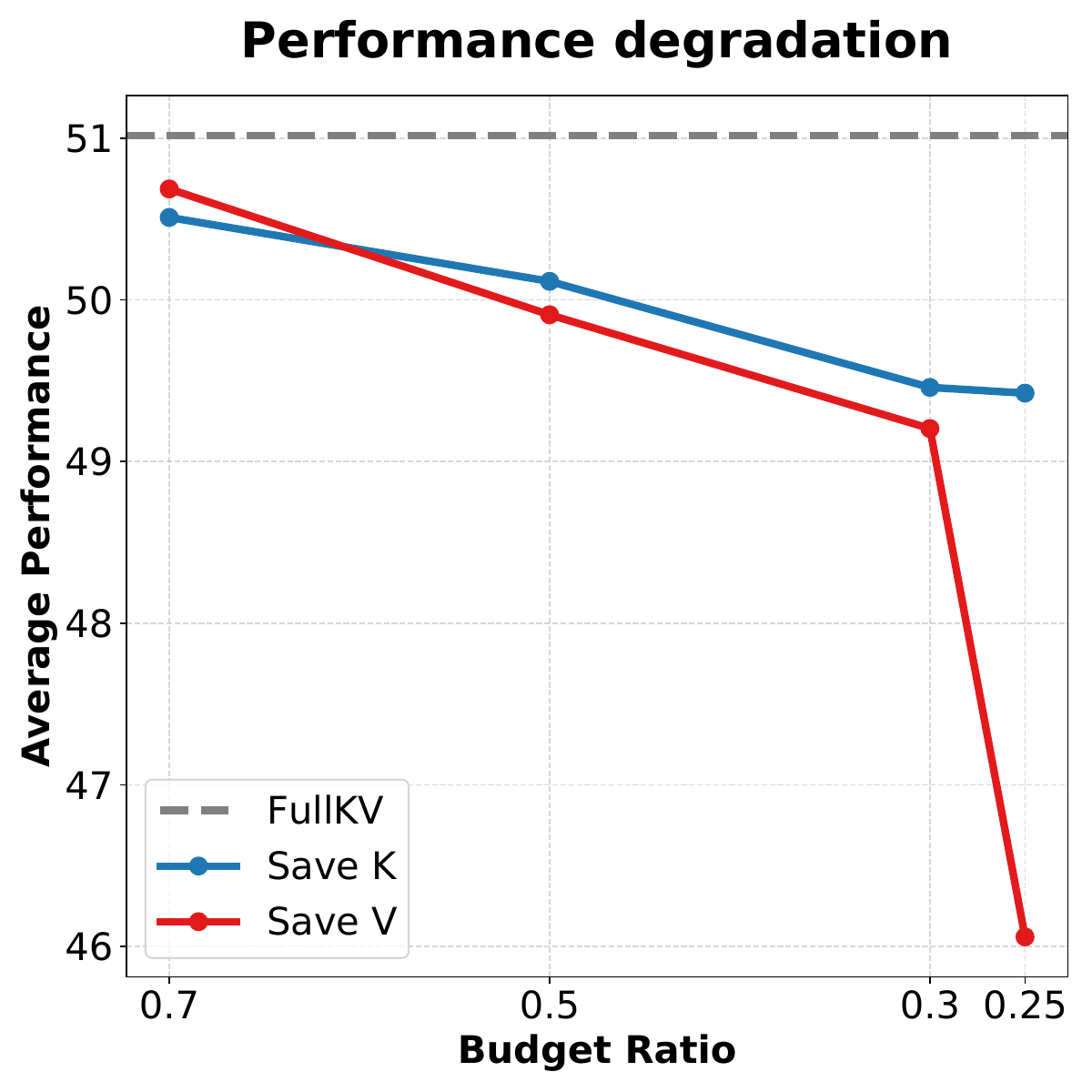}
   }
   \hfill
   \subcaptionbox{Performance and training time of the lightweight network across different training datasets.
   \label{fig:analy:dataset}}
   {\includegraphics[width=0.31\textwidth]{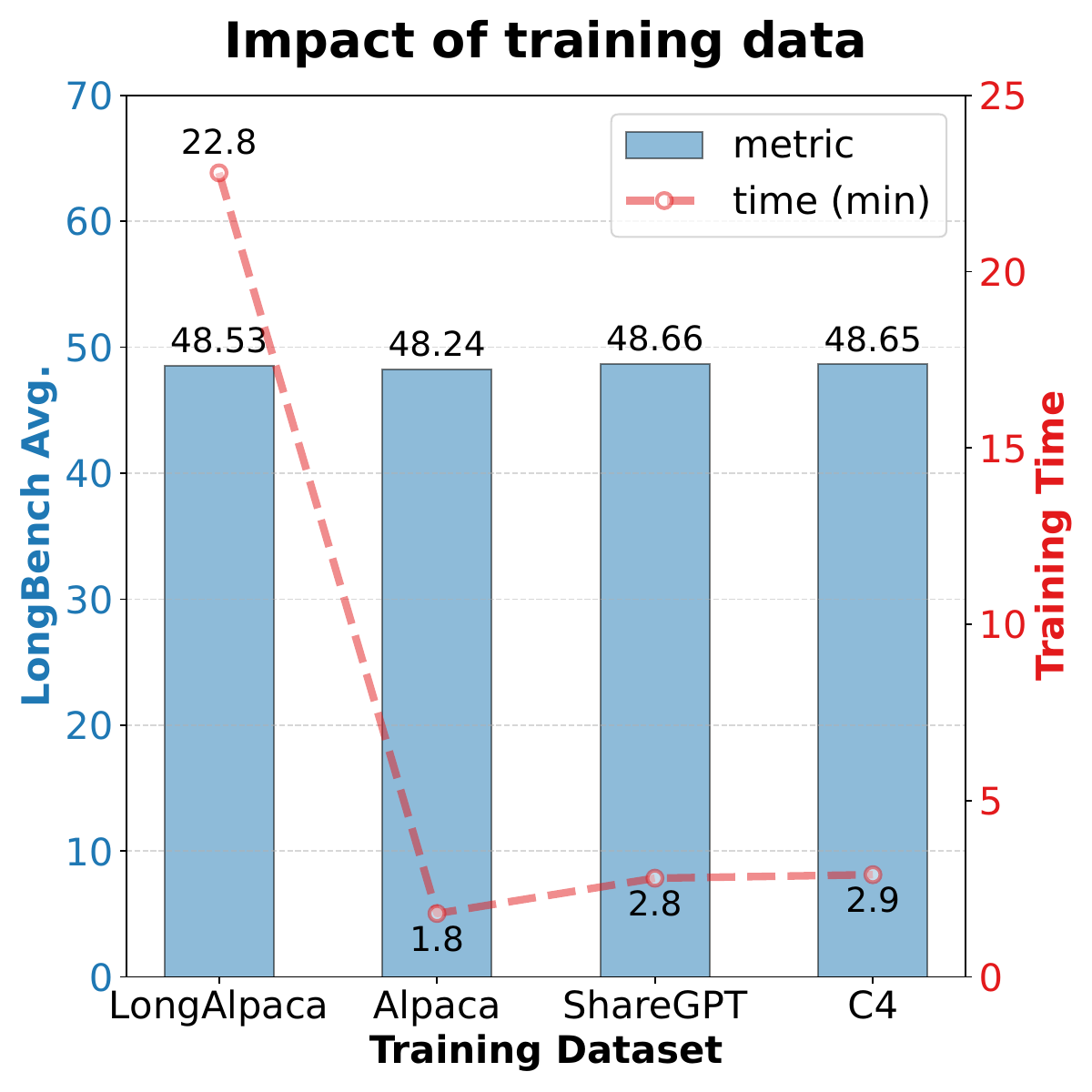}
   }
   \hfill
   \subcaptionbox{Throughput of FullKV, adaptive EchoKV, and static compression baselines across input lengths.
   \label{fig:analy:throughput}}
   {\includegraphics[width=0.31\textwidth]{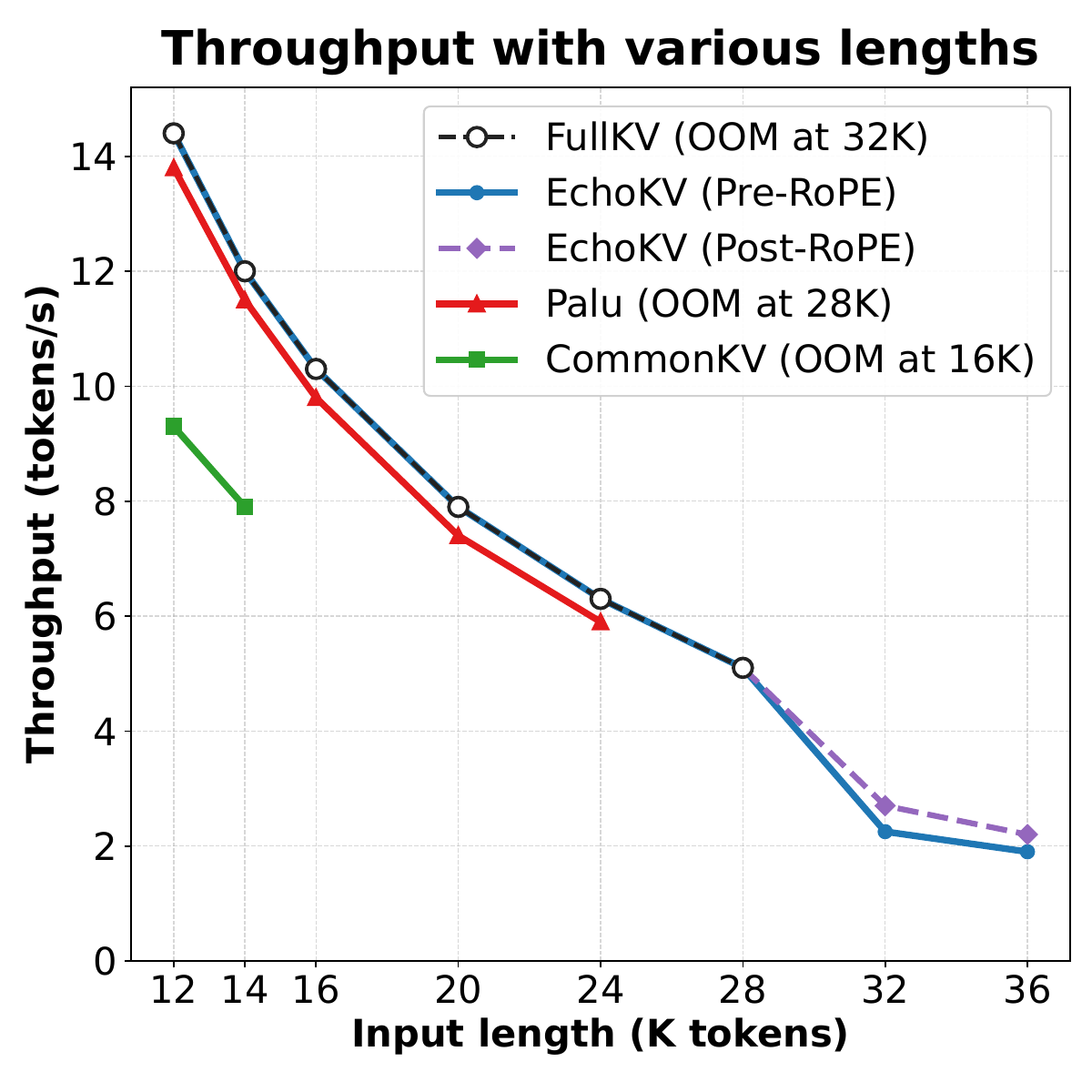}
   }
   \caption{
    Analysis experiments on EchoKV.
    All evaluations use Llama-3.1-8B-Instruct \citep{grattafiori2024llama}; (a,b) report LongBench \citep{bai2024longbench} results.
}
   \label{fig:analy}
\end{figure*}

\subsection{Asymmetric reconstruction difficulty of keys and values}
\label{sec:kv}
Recent studies \citep{liu2024kivi,cui2025homogeneous} indicate that keys and values exhibit distinct numerical distributions, necessitating separate treatment.
To compare the reconstruction difficulty of keys and values using EchoKV,
we conduct experiments where we reconstruct them separately.
As shown in Figure \ref{fig:analy:kv},
the ``Save K'' setting ensures key integrity by applying reconstruction exclusively to the values.
We evaluate the performance of the two configurations on LongBench across different budgets.
Experimental results indicate that compressing keys is more challenging than compressing values.
As the budget decreases, predicting only keys causes significant performance degradation, whereas predicting only values performs well.
This suggests that the inductive bias exploited by EchoKV, namely cross-layer and intra-layer similarity, is stronger for Values than for Keys.
Combined with prior observations that Keys are often more amenable to low-rank compression \citep{chang2024palu}, this asymmetry indicates that heterogeneous treatment of Keys and Values may be a promising direction for future KV compression methods.

\subsection{Impact of training data}
\label{sec:dataset}
Despite its high efficiency, EchoKV fundamentally requires training on a given dataset.
We experiment with different training datasets to verify robustness.
We test four datasets:
long instruction dataset LongAlpaca \citep{long-alpaca},
short instruction dataset Alpaca \citep{alpaca},
multi-turn instruction dataset ShareGPT 
\citep{sharegpt_chinese_english_90k}, 
and a subset of pre-training dataset C4 \citep{2019t5}.

As shown in Figure \ref{fig:analy:dataset},
we train the lightweight network on four distinct datasets and perform a unified evaluation on LongBench.
The resulting LongBench averages are highly comparable across these datasets,
even on the noisier pre-training subset.
This suggests that the lightweight predictor is primarily calibrating the model-specific similarity structure within the KV space, rather than relying strongly on the semantics of a particular training corpus.
In addition, while performance remains comparable, training durations vary substantially due to sequence length.
Even without sufficient long-context data, for Llama-3.1-8B we can train the lightweight network on short sequences in roughly 2 minutes on a single A100 GPU and still obtain strong long-context performance.

\begin{table*}[!t]
\centering
\caption{Results of the ablation study on different input features. 
Evaluations are conducted on LongBench using Llama-3.1-8B-Instruct with a compression ratio of 0.75.
\underline{\textbf{Bolded}} entries represent datasets where the two features exhibit differences.
}\label{tab:featrue}
\fontsize{20}{26}\selectfont
\setlength{\tabcolsep}{7pt}
\begin{threeparttable}
\scalebox{0.33}{
\begin{tabular}{cccccccccccccccccc}
\toprule[2pt]
\multirow{4}{*}{\textbf{~~~Methods}} & \multicolumn{3}{c}{\textbf{Single-Document QA}} & \multicolumn{3}{c}{\textbf{Multi-Document QA}}& \multicolumn{3}{c}{\textbf{Summarization}}& \multicolumn{3}{c}{\textbf{Few-shot Learning}}& \multicolumn{2}{c}{\textbf{Synthetic}} & \multicolumn{2}{c}{\textbf{Code}}
\\
\cmidrule(lr){2-4}\cmidrule(lr){5-7}\cmidrule(lr){8-10}\cmidrule(lr){11-13}\cmidrule(lr){14-15}\cmidrule(lr){16-17}
& \rotatebox[origin=c]{30}{\textbf{Nrtv}} & \rotatebox[origin=c]{30}{\textbf{Qasp}} & \rotatebox[origin=c]{30}{\textbf{MF}} & \rotatebox[origin=c]{30}{\textbf{Hotpot}} & \rotatebox[origin=c]{30}{\textbf{2Wiki}} & \rotatebox[origin=c]{30}{\textbf{Musiq}} & \rotatebox[origin=c]{30}{\textbf{Gov}} & \rotatebox[origin=c]{30}{\textbf{QMS}} & \rotatebox[origin=c]{30}{\textbf{MN}} & \rotatebox[origin=c]{30}{\textbf{TREC}} & \rotatebox[origin=c]{30}{\textbf{TQA}} & \rotatebox[origin=c]{30}{\textbf{SAM}} & \rotatebox[origin=c]{30}{\textbf{PC}} & \rotatebox[origin=c]{30}{\textbf{PR}} & \rotatebox[origin=c]{30}{\textbf{LCC}} & \rotatebox[origin=c]{30}{\textbf{RB}} \\
\midrule
~~~FullKV & 31.38 & 46.60 & 56.85 & 58.10 & 50.01 & 32.52 & \underline{\textbf{34.66}} & 25.19 & 27.11 & 73.00 & 92.11 & 43.70 & 6.56 & \underline{\textbf{100.00}} & 65.52 & 54.75 
\\

\hdashline

~~~Local Input & 31.88 & 46.45 & 57.32 & 58.02 & 49.49 & 32.68 & \underline{\textbf{30.80}} & 24.98 & 26.16 & 73.00 & 92.50 & 42.45 & 6.88 & \underline{\textbf{98.50}} & 64.72 & 55.53 
\\

~~~Global Input & 31.67 & 46.10 & 57.85 & 58.19 & 49.54 & 32.81 & \underline{\textbf{32.27}} & 24.63 & 26.32 & 73.00 & 92.41 & 41.77 & 7.00 & \underline{\textbf{88.00}} & 64.34 & 55.32 
\\

~~~Combined Input & 31.60 & 46.48 & 57.43 & 58.27 & 49.84 & 33.02 & \underline{\textbf{32.34}} & 25.09 & 26.22 & 73.00 & 92.58 & 42.49 & 7.00 & \underline{\textbf{94.50}} & 64.46 & 55.73
\\

\bottomrule[2pt]
\end{tabular}
}
\end{threeparttable}
\end{table*}

\subsection{Throughput analysis}
EchoKV supports on-demand compression as memory demand grows.
Figure \ref{fig:analy:throughput} compares adaptive EchoKV with FullKV and static compression baselines on a single NVIDIA A100-SXM4-80GB, with batch size 8, 16 output tokens, and a compression ratio of 0.5.
EchoKV uses the native FullKV path through 28K and activates compression at 32K/36K, where FullKV runs out of memory.
The Pre-RoPE and Post-RoPE variants reach 2.25/1.90 and 2.70/2.20 tokens/s at these lengths, respectively.
The quality benchmarks use compressed caches from prefill; the adaptive curves instead reflect when compression is enabled.

Appendix \ref{app:switching_results} evaluates length routing and memory-triggered conversion on 20 batches of 8 sequences, with prompts of 4K, 8K, 32K, and 36K.
Both adaptive policies complete all 20 batches, compared with 10 for FullKV, and retain comparable short-prompt time per output token (TPOT).
Memory-triggered conversion costs 193.52/165.52 ms at 32K/36K, followed by TPOT of 475.72/534.09 ms.
Cache reconstruction adds decoding work, trading latency for the capacity to complete longer requests.
Fusing reconstruction with cache assembly and writing directly into the attention buffer reduces intermediate materialization and redundant memory traffic.

\subsection{Ablation of input features}
As discussed in \S \ref{sec:arc}, the predictor input consists of the retained global cache from the first layer of the group and the partial local cache from the current layer.
We perform an ablation study to validate the effectiveness of various input features, 
with the results presented in Table \ref{tab:featrue}.
To ensure a fair evaluation, 
we fix the compression ratio at 0.75 and reconstruct the remaining cache using only different input features.
Experimental results indicate that global and local features achieve similar effectiveness across most datasets, demonstrating substantial redundancy both across layers and within the flattened per-layer KV representation.
However, in some summarization and synthetic tasks, 
the two types of features exhibit significant performance differences.
Fortunately, by combining these two features, we achieve consistent performance improvements.
We adopt this configuration in our main experiments, 
as it retains the lower-layer global cache to mitigate error accumulation within the group, 
while utilizing the local cache to ensure intra-layer precision.

\section{Conclusion}
In this paper, we introduced EchoKV, a novel and flexible KV cache compression framework designed to alleviate the memory bottleneck in long-context LLM inference.
By employing a lightweight network to predict the remaining cache segments from a subset, 
we achieve competitive performance across multiple benchmarks, with particular advantages at lower cache budgets.
We further report an exploratory hybrid variant in the appendix, indicating that differentiated compression mechanisms for Keys and Values are a promising direction beyond the main EchoKV design.

\begin{ack}
We gratefully acknowledge the support of the National Natural Science Foundation of China (NSFC) via grant 62236004 and 62476073.
\end{ack}

\bibliographystyle{unsrtnat}
\bibliography{custom}

\clearpage
\appendix

\section*{Appendices}
\noindent\textbf{A \quad Discussion and limitations} \dotfill \pageref{app:limitations}

\vspace{0.5em}
\noindent\textbf{B \quad Experimental details} \dotfill \pageref{app:details}\\[0.25em]
\hspace*{1.8em}B.1 \quad Hyperparameter configurations \dotfill \pageref{app:hyper_config}\\[0.2em]
\hspace*{1.8em}B.2 \quad Compression-ratio configurations \dotfill \pageref{app:config}\\[0.2em]
\hspace*{1.8em}B.3 \quad Training details \dotfill \pageref{app:hyperpara}\\[0.2em]
\hspace*{1.8em}B.4 \quad Inference and cache conversion \dotfill \pageref{app:inference}

\vspace{0.5em}
\noindent\textbf{C \quad Additional results} \dotfill \pageref{app:more_res}\\[0.25em]
\hspace*{1.8em}C.1 \quad Results on RULER \dotfill \pageref{app:ruler_results}\\[0.2em]
\hspace*{1.8em}C.2 \quad Comparison with TransMLA \dotfill \pageref{app:transmla_results}\\[0.2em]
\hspace*{1.8em}C.3 \quad Hybrid results \dotfill \pageref{app:hybrid_results}\\[0.2em]
\hspace*{1.8em}C.4 \quad Visualization of Needle-in-a-Haystack \dotfill \pageref{app:niah}\\[0.2em]
\hspace*{1.8em}C.5 \quad Adaptive inference and switching \dotfill \pageref{app:switching_results}\\[0.2em]
\hspace*{1.8em}C.6 \quad Ultra-long context \dotfill \pageref{app:ultralong}\\[0.2em]
\hspace*{1.8em}C.7 \quad Seed robustness \dotfill \pageref{app:seed_robustness}

\vspace{0.5em}
\noindent\textbf{D \quad Additional analysis} \dotfill \pageref{app:additional_analysis}\\[0.25em]
\hspace*{1.8em}D.1 \quad Analysis of QK-KL divergence loss \dotfill \pageref{app:qk_loss}\\[0.2em]
\hspace*{1.8em}D.2 \quad Group size, local dimension, and prediction distance \dotfill \pageref{app:sm_ablation}\\[0.2em]
\hspace*{1.8em}D.3 \quad Local-dimension selection \dotfill \pageref{app:local_select}\\[0.2em]
\hspace*{1.8em}D.4 \quad Prediction network architecture \dotfill \pageref{app:pred_arch}

\vspace{0.5em}
\noindent\textbf{E \quad Use of AI tools} \dotfill \pageref{app:ai}

\clearpage

\section{Discussion and limitations}
\label{app:limitations}
As discussed in Appendix \ref{app:local_select}, EchoKV adopts a simple prefix-based rule for retaining local dimensions, primarily for efficiency and contiguous memory access.
While this heuristic is already competitive, the results also suggest that more sophisticated selection strategies may still offer modest gains in some regimes.
Designing adaptive local-dimension selection mechanisms that better balance performance and implementation cost remains an open direction for future work.

We also find an intriguing asymmetry between Keys and Values.
Our analysis suggests that the EchoKV-style reconstruction mechanism is more effective for Values than for Keys, which contrasts with prior compression methods such as Palu \citep{chang2024palu} and XKV \citep{chang2025xkv}, where Keys are generally the easier component to compress.
This discrepancy indicates that Keys and Values may favor fundamentally different compression mechanisms, and a deeper study of their distributional and structural differences could lead to more principled hybrid designs.

Reconstructing and assembling K/V adds decoding latency, while tighter cache budgets can reduce task quality.
The evaluated conversion proceeds from FullKV to EchoKV; recovering discarded entries relies on reconstruction.
Serving measurements use the current PyTorch implementation on an A100-80GB, so latency and memory limits depend on the kernels and hardware.
The 64K/128K and multi-seed evaluations cover Llama-3.1-8B, with seed variation tested at ratios 0.7 and 0.5.

\section{Experimental details}
\label{app:details}

\subsection{Hyperparameter configurations}
\label{app:hyper_config}

Let $D=H_{\mathrm{KV}}d_{\mathrm{head}}$ be the flattened dimension of Keys or Values.
For each non-anchor layer ($S>1$), separate affine K/V predictors map $[B,L,D+m]$ to $[B,L,D-m]$, where $B$ is the batch size and $L$ is the number of tokens to reconstruct.
Each has a $(D-m)\times(D+m)$ weight matrix and a bias of length $D-m$.
At $m=0$, this becomes a $D\to D$ map from anchor to target.
Table \ref{tab:config_summary} lists the experimental configurations.

\begin{table*}[h]
    \centering
    \caption{EchoKV configurations for Llama-3.1-8B-Instruct, Mistral-7B-Instruct-v0.3, and Mistral-Small-24B-Instruct-2501. Predictor dimensions are shared across these models; approximate total K/V predictor counts refer to Llama-3.1-8B-Instruct and depend on the number of non-anchor layers.}
    \label{tab:config_summary}

    \begin{tabularx}{\linewidth}{
        >{\centering\arraybackslash}X 
        >{\centering\arraybackslash}X 
        >{\centering\arraybackslash}X 
        >{\centering\arraybackslash}X 
        >{\centering\arraybackslash}X 
        >{\centering\arraybackslash}X
    }
    \toprule
    \textbf{Target Ratio} & \textbf{Group Size ($S$)} & \textbf{Local Dim ($m$)} & \textbf{Predictor Input Dim} & \textbf{Predictor Output Dim} & \textbf{Total Params} \\
    \midrule
    0.7 & 2 & 384 & 1408 & 640 & 28.8M \\
    0.5 & 2 & 0 & 1024 & 1024 & 33.6M \\
    0.3 & 4 & 64 & 1088 & 960 & 50.2M \\
    \bottomrule
    \end{tabularx}
\end{table*}

\subsection{Compression-ratio configurations}
\label{app:config}

To achieve the target compression ratios, we choose the group size $S$ and retained local dimension $m$ using the budget construction in Appendix \ref{app:sm_ablation}.
In our experiments, Llama-3.1-8B-Instruct, Mistral-7B-Instruct-v0.3, and Mistral-Small-24B-Instruct-2501 all utilize Grouped Query Attention (GQA) with 8 KV heads and a head dimension of 128. Consequently, the flattened dimension of either Keys or Values is identical across these models, i.e., $D = 8 \times 128 = 1024$.
Based on this shared KV layout, the three models use the same predictor configurations in Table \ref{tab:config_summary}. For models with different KV dimensions, the same construction applies by adjusting the predictor input and output dimensions according to the target compression ratio.
For the three main configurations, the continuous local budgets $m^{*}=409.6$, $0$, and $68.27$ are rounded down to $m=384$, $0$, and $64$, respectively, using multiples of 64.

\paragraph{Compression ratio = 0.7.}
We set the group size to $S=2$.
In the compressed layer, we retain a local feature dimension of $m = 384$, corresponding to retaining 384 flattened local dimensions.
Consequently, the input to the lightweight linear network consists of the full global cache from the preceding layer ($1024$) concatenated with the local cache ($384$), resulting in an input dimension of $1408$.
The network predicts the remaining discarded features ($1024 - 384 = 640$).
Thus, the lightweight network $\mathbf{W}$ has a shape of $1408 \to 640$.
The nominal cache ratio for the compressed token region is $(1024 + 384) / (1024 \times 2) = 0.6875$.

\paragraph{Compression ratio = 0.5.}
We set the group size to $S=2$.
In this setting, we retain no local flattened dimensions ($m = 0$) for the compressed layer, relying entirely on the inter-layer similarity from the global input.
The linear layer maps the global input ($1024$) to the full cache of the current layer ($1024$).
The nominal cache ratio for the compressed token region is $(1024 + 0) / (1024 \times 2) = 0.5$.

\paragraph{Compression ratio = 0.3.}
We increase the group size to $S=4$.
For the three compressed layers within the group, we retain a minimal local feature dimension of $m = 64$, corresponding to retaining only 64 flattened local dimensions.
The lightweight network takes the global input ($1024$) and the small local embedding ($64$) to reconstruct the remaining information ($1024 - 64 = 960$).
The nominal cache ratio for the compressed token region is $(1024 + 64 \times 3) / (1024 \times 4) = 0.296875$.

A summary of these configurations is provided in Table \ref{tab:config_summary}.
These ratios describe complete layer groups over the compressed token region.
Full sink/window tokens and the decode tail contribute additional cache storage; Appendix \ref{app:inference} details the memory accounting.
In the batch-1 switching test at ratio 0.5, the measured persistent cache is 1.010/2.000 GiB at 16K and 2.010/4.000 GiB at 32K (EchoKV/FullKV), corresponding to 50.50\%/50.25\%.
The BF16 predictors add 0.0626 GiB, bringing cache-plus-predictor storage to 53.6\%/51.8\% of the FullKV cache; total peak GPU memory also includes the backbone and temporary tensors.

\subsection{Training details}
\label{app:hyperpara}

We implement EchoKV using the PyTorch framework. The lightweight reconstruction network is optimized using the \textbf{AdamW} optimizer. We employ a \textbf{Cosine} learning rate scheduler with no warmup steps to adjust the learning rate during training. To minimize memory overhead, the batch size is set to 1 for both stages.
For the main experiments, we set the global random seed to 42 for PyTorch and other relevant libraries; Appendix \ref{app:seed_robustness} additionally evaluates independently trained predictors with seeds 43--45.

The main training configuration uses two stages to progressively refine reconstruction quality:
\begin{itemize}
    \item \textbf{Stage 1 (Initialization):} The network is trained for 600 steps using the direct Key-Value reconstruction loss (KV-MSE) to establish a solid initialization.
    \item \textbf{Stage 2 (Adaptation):} We switch to the attention-aware output loss (O-MSE) and continue training for another 1,000 steps to align the reconstructed cache with the attention mechanism.
\end{itemize}

\begin{table}[t]
    \centering
    \captionsetup{skip=4pt}
    \caption{Hyperparameter details of the EchoKV training process. In Stage 2, only the parameters that differ from those in Stage 1 are described.}
    \label{tab:hyperparams}
    \begin{tabularx}{0.6\linewidth}{X c}
    \toprule
    \textbf{Configuration} & \textbf{Value} \\
    \midrule
    \multicolumn{2}{c}{\textbf{Stage 1}} \\
    \midrule
    Optimizer & AdamW \\
    Learning rate & $5\mathrm{e}{-4}$ \\
    Scheduler & Cosine Decay \\
    Warmup steps & 0 \\
    Batch size & 1 \\
    Training steps & 600 \\
    Loss function & KV-MSE \\
    \midrule
    \multicolumn{2}{c}{\textbf{Stage 2}} \\
    \midrule
     Optimizer & AdamW \\
    Learning rate & $5\mathrm{e}{-4}$ \\
    Scheduler & Cosine Decay \\
    Warmup steps & 0 \\
    Batch size & 1 \\
    Training steps & 1000 \\
    Loss function & O-MSE \\
    \bottomrule
    \end{tabularx}
\end{table}

Detailed hyperparameter configurations are provided in Table \ref{tab:hyperparams}.

\paragraph{Training schedules and objective ablation.}
The main LongAlpaca configuration takes 22.8 minutes on one A100; the shorter Alpaca, ShareGPT, and C4 settings in Figure \ref{fig:analy:dataset} train in a few minutes with comparable LongBench performance.
The objective ablation in Table \ref{tab:loss} instead uses 1,000 total steps: either a single objective throughout or 700 KV-MSE steps followed by 300 second-stage steps.
Table \ref{tab:loss_full} gives the full schedules and total workflow times, along with RULER averages over 13 tasks.
Compared with KV-MSE alone, two-stage O-MSE improves LongBench/RULER by 0.27/1.70 points with two additional minutes of training.

\begin{table}[htbp]
    \centering
    \caption{Complete objective-ablation results. Times cover the entire training workflow; schedules list the steps for each objective in order.}
    \label{tab:loss_full}
    \small
    \begin{tabular}{lcccc}
    \toprule
    \textbf{Training objective} & \textbf{Steps} & \textbf{Total time} & \textbf{LongBench} & \textbf{RULER} \\
    \midrule
    KV-MSE only & 1,000 & 12.5 min & 48.99 & 81.23 \\
    O-MSE only & 1,000 & 15.4 min & 48.76 & 78.74 \\
    KV-MSE then QK-KL & $700+300$ & 39.7 min & 49.11 & 81.33 \\
    KV-MSE then O-MSE & $700+300$ & 14.5 min & \textbf{49.26} & \textbf{82.93} \\
    \bottomrule
    \end{tabular}
\end{table}

\subsection{Inference and cache conversion}
\label{app:inference}

\paragraph{Prefill and decoding.}
Prefill attention uses full K/V, while the cache is stored directly in EchoKV form.
Each group retains a full anchor cache; other layers retain full K/V for 4 sink tokens and a 128-token local window, plus the first $m$ flattened dimensions for intervening tokens.
Decoding reconstructs this middle region from the anchor and local features, then assembles it with the retained full K/V for attention.
Newly generated tokens are appended in full to the cache tail.

\paragraph{Conversion from FullKV.}
Inference starts on the native FullKV path.
A trigger based on generation length or memory usage compacts the existing cache to a preselected $(S,m)$ configuration, retaining the anchors, sink/window tokens, and local dimensions.
Decoding continues with the corresponding trained predictors, reusing the cached prefix.

\paragraph{RoPE handling.}
The main Pre-RoPE configuration reconstructs unrotated Keys and applies RoPE at their original positions before attention.
Conversion from native FullKV applies inverse RoPE before compaction.
The Post-RoPE variant reconstructs rotated Keys, saving the inverse transformation at switching and repeated rotation during decoding.

\paragraph{Memory accounting.}
Persistent KV storage comprises anchors, full sink/window caches and the decode tail, and retained local dimensions.
Predictor parameters are counted separately; reconstructed K/V form per-layer workspace.
Total peak GPU memory also includes backbone parameters, activations, and temporary tensors for conversion and attention.

\paragraph{Mid-session switching protocol.}
We evaluate Llama-3.1-8B-Instruct at ratio 0.5 with greedy decoding.
Switch@$t$ generates $t$ tokens with FullKV, converts the populated cache, and generates token $t+1$ onward with EchoKV.
Quality is evaluated at $t\in\{32,64\}$ on GovReport, QMSum, and MultiNews using the LongBench prompts and generation limits.
Timing uses Switch@32 with 16K/32K prompts, 256 output tokens, and batch size 1 on an A100-80GB.
Latencies are averaged over five warmed-up runs.
Switch latency includes inverse RoPE for Pre-RoPE; time per output token (TPOT) measures steady decoding after conversion.

\paragraph{Adaptive workload.}
Using the same model and ratio with Pre-RoPE, we evaluate 20 batches of 8 sequences (160 sequences): five batches each at 4K, 8K, 32K, and 36K, with 16 output tokens and the same shuffled order for every policy.
The length policy routes prompts of at least 32K to EchoKV before prefill.
The memory policy starts with FullKV and converts completed layers at a 62-GiB peak-memory threshold during prefill, then continues with EchoKV.
Appendix \ref{app:switching_results} reports routing performance and switching quality.

\section{Additional results}
\label{app:more_res}
\subsection{Results on RULER}
\label{app:ruler_results}
In this section, we present the detailed experimental results on the RULER benchmark.
Table \ref{tab:ruler_full} summarizes the performance of Llama-3.1-8B-Instruct, Mistral-7B-Instruct-v0.3, and Mistral-Small-24B-Instruct-2501 across varying compression ratios (0.5 and 0.3).

\begin{table*}[t]
\fontsize{20}{26}\selectfont
\setlength{\tabcolsep}{12pt}
\centering
\caption{
Experimental results of different methods on RULER at compression ratios of 0.5 and 0.3.
$^*$ Indicates the actual compression ratio is 0.6.
}\label{tab:ruler_full}
\begin{threeparttable}
\scalebox{0.345}{
\begin{tabular}{lcccccccccccccc}
\toprule[2pt]
\multirow{4}{*}{\textbf{Method}} & \multicolumn{8}{c}{\textbf{Needle-in-a-Haystack (NIAH)}} & \multicolumn{3}{c}{\textbf{Synthetic}}
& \multicolumn{2}{c}{\textbf{QA}}
&\multirow{4}{*}{\textbf{~~~Avg.}} \\
\cmidrule(lr){2-9}\cmidrule(lr){10-12}\cmidrule(lr){13-14}
& \rotatebox[origin=c]{30}{\textbf{S1}} & \rotatebox[origin=c]{30}{\textbf{S2}} & \rotatebox[origin=c]{30}{\textbf{S3}} & \rotatebox[origin=c]{30}{\textbf{MK1}} & \rotatebox[origin=c]{30}{\textbf{MK2}} & \rotatebox[origin=c]{30}{\textbf{MK3}} & \rotatebox[origin=c]{30}{\textbf{MV}} & \rotatebox[origin=c]{30}{\textbf{MQ}} & \rotatebox[origin=c]{30}{\textbf{VT}} & \rotatebox[origin=c]{30}{\textbf{CWE}} & \rotatebox[origin=c]{30}{\textbf{FWE}} & \rotatebox[origin=c]{30}{\textbf{QA1}} & \rotatebox[origin=c]{30}{\textbf{QA2}} & \\
\midrule

\multicolumn{15}{c}{\textbf{Llama-3.1-8B-Instruct}} \\
\midrule

\multicolumn{15}{c}{\cellcolor{lightgray!25} \textit{Compression Ratio = 0.5}} \\
FullKV & 100.0 & 100.0 & 100.0 & 99.00 & 99.00 & 99.00 & 98.75 & 98.50 & 99.20 & 17.80 & 87.00 & 87.00 & 54.00 & 87.63 \\
\hdashline
ThinK* & 1.00 & 2.00 & 0.00 & 2.00 & 0.00 & 0.00 & 1.75 & 3.25 & 0.60 & 0.00 & 48.00 & 63.00 & 40.00 & 12.43 \\
Palu & \textbf{100.0} & 98.00 & 75.00 & 97.00 & 78.00 & 30.00 & 73.00 & 64.25 & 86.00 & \textbf{15.30} & 78.33 & 53.00 & 38.00 & 68.14 \\
CommonKV & 92.00 & 94.00 & 83.00 & 97.00 & 67.00 & 1.00 & 87.00 & 94.25 & 79.80 & 0.10 & \textbf{85.67} & 75.00 & 50.00 & 69.68 \\
EchoKV & \textbf{100.0} & \textbf{100.0} & \textbf{100.0} & \textbf{99.00} & \textbf{99.00} & 94.00 & 89.50 & 99.50 & 87.20 & 0.60 & 80.00 & \textbf{85.00} & 52.00 & 83.52 \\
EchoKV-Hybrid & \textbf{100.0} & \textbf{100.0} & \textbf{100.0} & \textbf{99.00} & \textbf{99.00} & \textbf{99.00} & \textbf{97.00} & \textbf{100.0} & \textbf{97.40} & 13.80 & 79.67 & \textbf{85.00} & \textbf{54.00} & \textbf{86.45} \\

\multicolumn{15}{c}{\cellcolor{lightgray!25} \textit{Compression Ratio = 0.3}} \\
Palu & 11.00 & 0.00 & 0.00 & 0.00 & 0.00 & 0.00 & 0.00 & 0.00 & 0.00 & 0.30 & 1.00 & 0.00 & 0.00 & 0.95 \\
CommonKV & 45.00 & 28.00 & 0.00 & 13.00 & 0.00 & 0.00 & 7.50 & 8.50 & \textbf{38.40} & \textbf{9.90} & 39.33 & 21.00 & 30.00 & 18.51 \\
EchoKV & \textbf{83.00} & \textbf{81.00} & \textbf{72.00} & 70.00 & \textbf{79.00} & 9.00 & 57.25 & 88.00 & 37.00 & 0.00 & \textbf{75.67} & \textbf{82.00} & \textbf{50.00} & \textbf{60.30} \\
EchoKV-Hybrid & 57.00 & 79.00 & 46.00 & \textbf{88.00} & 78.00 & \textbf{17.00} & \textbf{71.25} & \textbf{93.50} & 31.40 & 0.10 & 67.67 & \textbf{82.00} & \textbf{50.00} & 58.53 \\

\midrule

\multicolumn{15}{c}{\textbf{Mistral-7B-Instruct-v0.3}} \\
\midrule

\multicolumn{15}{c}{\cellcolor{lightgray!25} \textit{Compression Ratio = 0.5}} \\
FullKV & 100.0 & 92.00 & 100.0 & 84.00 & 80.00 & 57.00 & 92.25 & 92.25 & 95.80 & 80.70 & 89.67 & 65.00 & 47.00 & 82.74 \\
\hdashline
ThinK* & 21.00 & 58.00 & 2.00 & 32.00 & 17.00 & 5.00 & 20.25 & 14.00 & 44.20 & 17.10 & 89.67 & 60.00 & 42.00 & 32.47 \\
Palu & \textbf{100.0} & \textbf{97.00} & 98.00 & 76.00 & 51.00 & 26.00 & 88.50 & 80.75 & 83.00 & 32.10 & 85.00 & 41.00 & 33.00 & 68.56 \\
CommonKV & 86.00 & 46.00 & 36.00 & 27.00 & 18.00 & 0.00 & 7.00 & 12.75 & 66.60 & 15.10 & 75.00 & 46.00 & 44.00 & 36.88 \\
EchoKV & \textbf{100.0} & 88.00 & 99.00 & 82.00 & 59.00 & 19.00 & 62.75 & 67.50 & 85.20 & 46.60 & 81.67 & 65.00 & 43.00 & 69.13 \\
EchoKV-Hybrid & \textbf{100.0} & 89.00 & \textbf{100.0} & \textbf{83.00} & \textbf{77.00} & \textbf{42.00} & \textbf{92.50} & \textbf{90.25} & \textbf{90.80} & \textbf{66.30} & \textbf{93.67} & \textbf{67.00} & \textbf{47.00} & \textbf{79.89} \\

\multicolumn{15}{c}{\cellcolor{lightgray!25} \textit{Compression Ratio = 0.3}} \\
Palu & 40.00 & 27.00 & 0.00 & 4.00 & 0.00 & 0.00 & 1.75 & 1.00 & 8.40 & 4.30 & 37.67 & 12.00 & 24.00 & 12.32 \\
CommonKV & 11.00 & 0.00 & 0.00 & 2.00 & 0.00 & 0.00 & 0.00 & 1.25 & 1.60 & 12.30 & 27.00 & 15.00 & 25.00 & 7.32 \\
EchoKV & 77.00 & 64.00 & 28.00 & 54.00 & 14.00 & 1.00 & 6.00 & 9.00 & 39.80 & 15.30 & 51.00 & 56.00 & 41.00 & 35.08 \\
EchoKV-Hybrid & \textbf{99.00} & \textbf{80.00} & \textbf{88.00} & \textbf{76.00} & \textbf{53.00} & \textbf{10.00} & \textbf{48.00} & \textbf{50.75} & \textbf{80.00} & \textbf{27.10} & \textbf{82.33} & \textbf{61.00} & \textbf{45.00} & \textbf{61.55} \\

\midrule

\multicolumn{15}{c}{\textbf{Mistral-Small-24B-Instruct-2501}} \\
\midrule

\multicolumn{15}{c}{\cellcolor{lightgray!25} \textit{Compression Ratio = 0.5}} \\
FullKV & 98.00 & 100.0 & 100.0 & 95.00 & 98.00 & 94.00 & 99.75 & 99.00 & 100.0 & 95.90 & 92.00 & 83.00 & 63.00 & 93.67 \\
\hdashline
ThinK* & 0.00 & 1.00 & 1.00 & 1.00 & 0.00 & 0.00 & 2.75 & 2.50 & 0.40 & 12.80 & 25.33 & 59.00 & 46.00 & 11.68 \\
Palu & 98.00 & \textbf{100.0} & 68.00 & \textbf{97.00} & 84.00 & 76.00 & 94.50 & 96.25 & 71.20 & 34.20 & 80.00 & 55.00 & 58.00 & 77.86 \\
CommonKV & 99.00 & 99.00 & 98.00 & 98.00 & 92.00 & 90.00 & 94.50 & \textbf{98.00} & 89.60 & 71.10 & \textbf{94.00} & 81.00 & \textbf{62.00} & 89.71 \\
EchoKV & 99.00 & \textbf{100.0} & \textbf{100.0} & 95.00 & \textbf{98.00} & \textbf{95.00} & 85.00 & 96.25 & \textbf{98.00} & 75.20 & 92.33 & 83.00 & 60.00 & 90.52 \\
EchoKV-Hybrid & \textbf{100.0} & \textbf{100.0} & \textbf{100.0} & 94.00 & 97.00 & 94.00 & 97.50 & 96.75 & 97.20 & \textbf{75.50} & 92.00 & \textbf{84.00} & 60.00 & \textbf{91.38} \\

\multicolumn{15}{c}{\cellcolor{lightgray!25} \textit{Compression Ratio = 0.3}} \\
Palu & 0.00 & 0.00 & 0.00 & 0.00 & 0.00 & 0.00 & 0.00 & 0.00 & 0.00 & 0.00 & 0.00 & 4.00 & 4.00 & 0.62 \\
CommonKV & 4.00 & 36.00 & 4.00 & 6.00 & 1.00 & 0.00 & 7.00 & 0.75 & 0.00 & 5.60 & 32.33 & 30.00 & 39.00 & 12.74 \\
EchoKV & \textbf{99.00} & \textbf{95.00} & \textbf{84.00} & \textbf{90.00} & \textbf{96.00} & \textbf{6.00} & \textbf{72.25} & \textbf{91.25} & \textbf{93.60} & \textbf{46.40} & \textbf{90.33} & \textbf{80.00} & \textbf{59.00} & \textbf{77.14} \\
EchoKV-Hybrid & 5.00 & 4.00 & 1.00 & 3.00 & 7.00 & 0.00 & 0.00 & 2.75 & 9.80 & 17.20 & 61.00 & 61.00 & 47.00 & 16.83 \\

\bottomrule[2pt]
\end{tabular}
}
\end{threeparttable}
\end{table*}

\subsection{Comparison with TransMLA}
\label{app:transmla_results}
We further compare EchoKV with TransMLA \citep{meng2025transmla}, a post-training method that converts existing GQA attention modules into an MLA-style architecture.
For fairness, we report two TransMLA settings on LongBench: direct conversion without additional adaptation (``Converted''), and a continued-training variant using the same LongAlpaca data as EchoKV (``FT'').
For both Llama-3.1-8B and Mistral-7B, we use the same conversion parameters: at a compression ratio of 0.5, we set \texttt{qk-mqa-dim} to 128 and \texttt{kv-lora-rank} to 896; at a compression ratio of 0.3, we set \texttt{qk-mqa-dim} to 128 and \texttt{kv-lora-rank} to 512.

Table \ref{tab:transmla} in the main text summarizes the average LongBench results.
Direct conversion leads to severe quality degradation on both backbones, while continued training substantially recovers performance.
However, even after this additional adaptation, EchoKV still maintains a clear advantage across both compression ratios on Llama-3.1-8B and Mistral-7B.

\begin{figure}[t]
    \centering
    \includegraphics[width=0.95\linewidth]{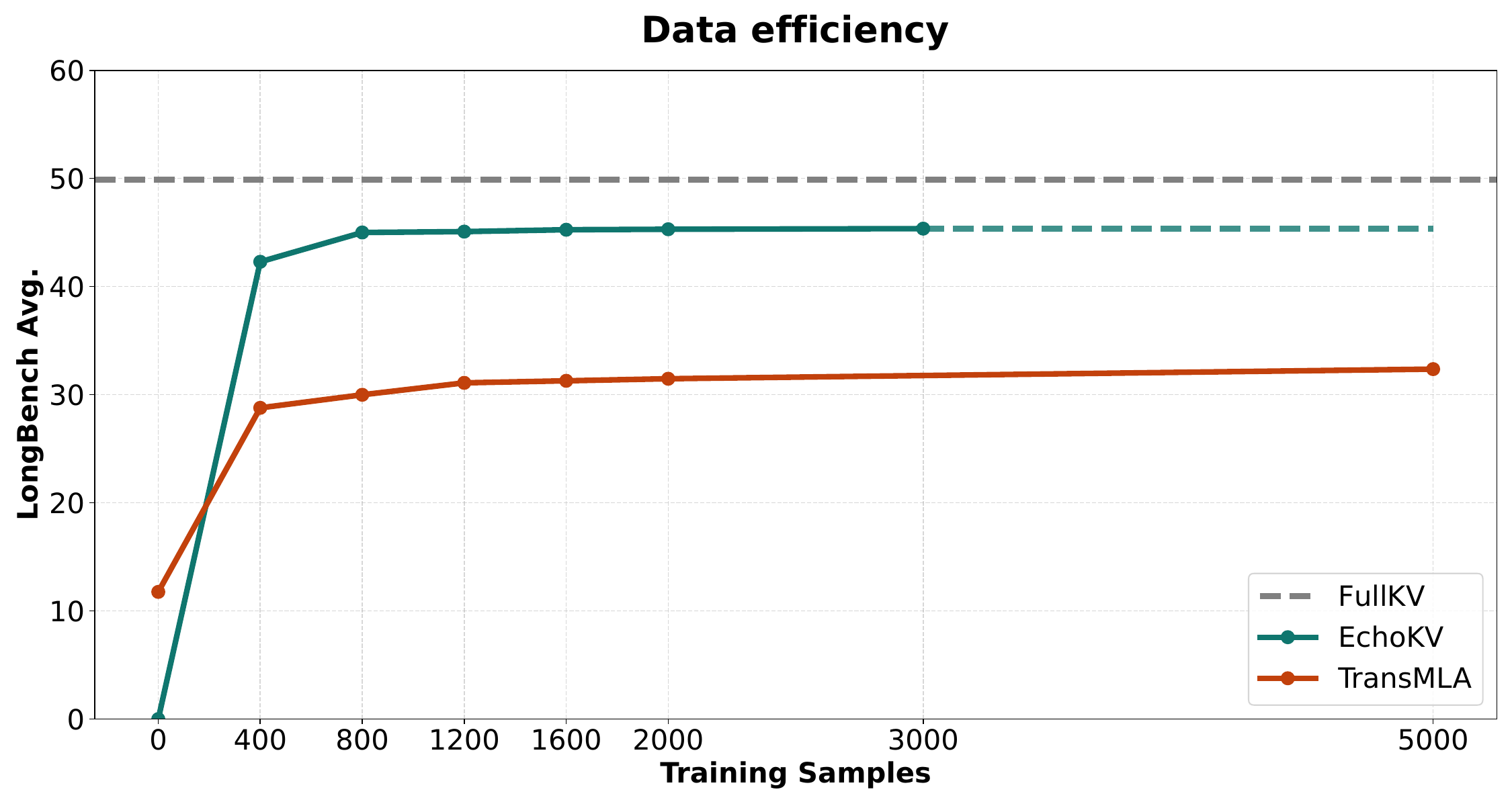}
    \caption{LongBench average as a function of training data size on Llama-3.1-8B at a compression ratio of 0.3. EchoKV reaches strong performance with substantially fewer training samples than TransMLA.}
    \label{fig:transmla_data_scaling}
\end{figure}

Figure \ref{fig:transmla_data_scaling} further compares data efficiency on Llama-3.1-8B at a compression ratio of 0.3.
EchoKV improves rapidly with a small amount of data and saturates early, whereas TransMLA starts from a much weaker converted model and requires substantially more training data to recover.
This comparison highlights a practical distinction between the two approaches: TransMLA must first restore a modified attention module after architecture conversion, whereas EchoKV preserves the original attention projections and trains only a lightweight reconstruction module.

\subsection{Hybrid results}
\label{app:hybrid_results}
We additionally report the results of EchoKV-Hybrid in the appendix, since this variant is intended as an exploratory extension rather than the main method.
The motivation comes from the asymmetric reconstruction difficulty analyzed in \S\ref{sec:kv}: Values are generally easier to reconstruct with EchoKV, whereas Keys are more sensitive to prediction errors.
Combined with prior observations that Keys are often more amenable to low-rank compression \citep{chang2024palu}, this suggests a heterogeneous design in which Keys and Values need not share the same compression mechanism.
Accordingly, EchoKV-Hybrid applies ThinK-style low-rank compression to Keys and EchoKV reconstruction to Values.

The results in Tables \ref{tab:ruler_full} and \ref{tab:longbench_hybrid} support this motivation, but also reveal its limits.
At moderate compression ratios, the hybrid design often provides additional gains over plain EchoKV, especially on several LongBench settings and on the RULER evaluations of Llama-3.1-8B-Instruct and Mistral-7B-Instruct-v0.3.
These improvements are consistent with the hypothesis that the inter-cache similarity exploited by EchoKV is more pronounced for Values, while Keys can benefit more from direct low-rank compression.
At the same time, the hybrid variant is not uniformly superior: under more aggressive compression, and particularly on the stronger 24B backbone, plain EchoKV is often the more robust choice.
Taken together, these results suggest that differentiated treatment of Keys and Values is promising, but that the best hybrid strategy likely depends on both the compression regime and the backbone model.

\begin{table*}[t]
\fontsize{22}{26}\selectfont
\centering
\caption{Experimental results on LongBench including the exploratory EchoKV-Hybrid variant.
$^*$ Since ThinK only compresses Keys, we double the compression rate calculation to account for Value storage. We report results at 0.6 because the 0.5 ratio is infeasible.
Compression Ratio = Size of Compressed Cache / Size of Full Cache.
}
\label{tab:longbench_hybrid}
\begin{threeparttable}
\setlength{\tabcolsep}{8pt}
\resizebox{\linewidth}{!}{%
\begin{tabular}{llccccccccccccccccc}
\toprule[2pt]
&\multirow{4}{*}{\textbf{~~~Methods}} & \multicolumn{3}{c}{\textbf{Single-Document QA}} & \multicolumn{3}{c}{\textbf{Multi-Document QA}}& \multicolumn{3}{c}{\textbf{Summarization}}& \multicolumn{3}{c}{\textbf{Few-shot Learning}}& \multicolumn{2}{c}{\textbf{Synthetic}} & \multicolumn{2}{c}{\textbf{Code}}&\multirow{4}{*}{\textbf{~~~Avg.}} \\
\cmidrule(lr){3-5}\cmidrule(lr){6-8}\cmidrule(lr){9-11}\cmidrule(lr){12-14}\cmidrule(lr){15-16}\cmidrule(lr){17-18}
&& \rotatebox[origin=c]{30}{\textbf{Nrtv}} & \rotatebox[origin=c]{30}{\textbf{Qasp}} & \rotatebox[origin=c]{30}{\textbf{MF}} & \rotatebox[origin=c]{30}{\textbf{Hotpot}} & \rotatebox[origin=c]{30}{\textbf{2Wiki}} & \rotatebox[origin=c]{30}{\textbf{Musiq}} & \rotatebox[origin=c]{30}{\textbf{Gov}} & \rotatebox[origin=c]{30}{\textbf{QMS}} & \rotatebox[origin=c]{30}{\textbf{MN}} & \rotatebox[origin=c]{30}{\textbf{TREC}} & \rotatebox[origin=c]{30}{\textbf{TQA}} & \rotatebox[origin=c]{30}{\textbf{SAM}} & \rotatebox[origin=c]{30}{\textbf{PC}} & \rotatebox[origin=c]{30}{\textbf{PR}} & \rotatebox[origin=c]{30}{\textbf{LCC}} & \rotatebox[origin=c]{30}{\textbf{RB}} \\

\midrule
\multirow{19}{*}{\rotatebox[origin=c]{90}{\fontsize{18}{100}\selectfont \textbf{Llama-3.1-8B-Instruct}}}

&~~~FullKV & 31.38 & 46.60 & 56.85 & 58.10 & 50.01 & 32.52 & 34.66 & 25.19 & 27.11 & 73.00 & 92.11 & 43.70 & 6.56 & 100.0 & 65.52 & 54.75 & 49.88 \\
\cline{2-19}

& \multicolumn{18}{c}{\cellcolor{lightgray!25} 
  \textit{Compression Ratio = 0.7}
} \\

&~~~MiniCache & 17.87 & 22.26 & 29.51 & 26.72 & 11.47 & 9.71 & 24.61 & 21.85 & 23.29 & 65.00 & 75.05 & 31.46 & 5.83 & 60.50 & 41.34 & 37.19 & 31.48 \\

&~~~ThinK & \textbf{32.77} & 45.44 & \textbf{58.83} & \textbf{58.30} & \textbf{51.09} & \textbf{32.65} & 32.63 & 24.96 & 26.19 & \textbf{73.00} & 91.72 & 42.10 & \textbf{6.96} & 99.50 & \textbf{65.94} & 55.64 & \textbf{49.86} \\

&~~~Palu & 28.96 & \textbf{46.73} & 53.26 & 53.08 & 45.35 & 29.91 & 32.51 & 24.69 & 25.42 & \textbf{73.00} & 89.10 & 43.77 & 2.50 & 96.50 & 57.90 & \textbf{56.78} & 47.47 \\

&~~~CommonKV & 29.87 & 41.16 & 53.86 & 54.90 & 45.42 & 29.96 & 27.91 & 24.21 & 25.68 & 66.00 & 90.03 & 43.10 & 6.67 & \textbf{100.0} & 61.92 & 53.41 & 47.13 \\

&~~~EchoKV & 32.24 & 44.53 & 57.84 & 57.90 & 49.89 & 32.08 & 29.29 & 25.06 & 25.02 & \textbf{73.00} & \textbf{92.52} & 41.67 & 6.87 & 99.00 & 63.57 & 54.72 & 49.08 \\

&~~~EchoKV-Hybrid & 32.04 & 45.97 & 58.15 & 58.14 & 50.16 & 32.42 & \textbf{33.27} & \textbf{25.12} & \textbf{26.82} & \textbf{73.00} & 92.28 & \textbf{43.80} & 6.27 & 97.00 & 65.12 & 55.13 & 49.67 \\

\cline{2-19}
& \multicolumn{18}{c}{\cellcolor{lightgray!25} 
  \textit{Compression Ratio = 0.5}
} \\

&~~~ThinK* & 28.51 & 22.92 & 37.42 & 52.90 & 47.84 & 28.01 & 19.67 & 21.96 & 19.01 & 38.00 & 88.11 & 40.72 & 7.50 & \textbf{99.50} & 59.97 & 49.92 & 41.37 \\

&~~~Palu & 19.27 & 34.64 & 40.14 & 41.99 & 26.71 & 22.45 & 22.60 & 23.70 & 23.21 & 66.00 & 25.95 & 39.75 & 4.00 & 49.50 & 29.86 & 30.19 & 31.25 \\

&~~~CommonKV & 25.24 & 40.64 & 54.45 & 54.56 & 45.58 & 29.82 & 25.23 & 23.85 & 24.92 & 59.00 & 89.98 & \textbf{42.50} & \textbf{10.12} & \textbf{99.50} & 62.46 & 52.50 & 46.27 \\

&~~~EchoKV & \textbf{33.06} & 44.80 & 57.51 & 57.91 & \textbf{50.25} & 31.79 & 29.22 & 24.10 & 25.21 & \textbf{73.00} & \textbf{91.92} & 41.45 & 7.17 & 89.50 & 64.61 & 54.91 & 48.53 \\

&~~~EchoKV-Hybrid & 31.93 & \textbf{45.44} & \textbf{57.68} & \textbf{58.53} & 50.13 & \textbf{32.69} & \textbf{31.81} & \textbf{24.81} & \textbf{25.83} & \textbf{73.00} & 91.87 & 42.48 & 6.92 & 96.50 & \textbf{65.74} & \textbf{55.11} & \textbf{49.40} \\

\cline{2-19}
& \multicolumn{18}{c}{\cellcolor{lightgray!25} 
  \textit{Compression Ratio = 0.3}
} \\
&~~~Palu & 1.90 & 2.42 & 4.36 & 1.00 & 1.70 & 0.66 & 3.52 & 9.28 & 5.95 & 1.25 & 3.19 & 6.31 & 0.00 & 0.00 & 18.20 & 18.79 & 4.91 \\

&~~~CommonKV & 11.20 & 29.58 & 36.89 & 31.21 & 31.30 & 14.79 & 13.86 & 21.22 & 21.82 & 47.50 & 83.38 & 40.25 & 0.22 & 8.50 & 54.43 & 51.10 & 31.08 \\

&~~~EchoKV & \textbf{31.22} & \textbf{42.47} & \textbf{54.27} & 57.03 & \textbf{49.58} & \textbf{30.75} & \textbf{24.23} & \textbf{23.84} & \textbf{22.32} & 49.00 & \textbf{91.79} & 40.90 & \textbf{7.25} & 84.50 & 62.09 & \textbf{53.03} & 45.27 \\

&~~~EchoKV-Hybrid & 30.89 & 38.41 & 50.83 & \textbf{57.28} & 49.07 & 30.72 & 23.21 & 23.25 & 21.39 & \textbf{57.50} & 90.85 & \textbf{41.77} & 6.77 & \textbf{95.50} & \textbf{62.77} & 51.69 & \textbf{45.74} \\

\midrule

\multirow{19}{*}{\rotatebox[origin=c]{90}{\fontsize{18}{100}\selectfont \textbf{Mistral-7B-Instruct-v0.3}}}

&~~~FullKV & 29.80 & 39.12 & 50.45 & 50.28 & 36.31 & 26.45 & 34.11 & 25.93 & 26.56 & 76.00 & 88.89 & 46.97 & 5.50 & 97.00 & 61.47 & 62.57 & 47.34 \\
\cline{2-19}

& \multicolumn{18}{c}{\cellcolor{lightgray!25} 
  \textit{Compression Ratio = 0.7}
} \\

&~~~MiniCache & 14.49 & 21.16 & 26.26 & 18.38 & 18.16 & 6.33 & 19.12 & 21.06 & 23.22 & 61.00 & 82.68 & 32.60 & 2.67 & 24.17 & 44.28 & 40.75 & 28.52 \\

&~~~ThinK & \textbf{30.42} & 37.87 & 50.13 & 50.35 & 34.55 & 25.92 & \textbf{33.81} & \textbf{25.66} & 26.40 & \textbf{76.00} & 88.43 & 45.81 & \textbf{5.50} & 95.00 & 61.90 & 61.95 & 46.86 \\

&~~~Palu & 28.47 & 36.08 & \textbf{52.56} & 46.57 & \textbf{36.12} & 26.37 & 33.59 & 25.16 & \textbf{26.54} & 73.50 & 87.77 & 45.11 & 4.00 & 94.50 & 59.04 & 61.06 & 46.03 \\

&~~~CommonKV & 26.02 & 37.13 & 48.21 & 47.61 & 29.87 & 24.43 & 30.05 & 24.75 & 25.76 & 74.00 & 88.12 & 44.54 & 4.00 & 92.50 & 61.16 & 62.10 & 45.02 \\

&~~~EchoKV & 29.93 & 37.65 & 49.43 & \textbf{51.50} & 34.54 & 24.98 & 32.29 & 25.11 & 25.99 & \textbf{76.00} & \textbf{89.47} & \textbf{47.44} & \textbf{5.50} & \textbf{95.50} & \textbf{62.26} & 62.18 & 46.86 \\

&~~~EchoKV-Hybrid & 30.24 & \textbf{38.90} & 50.27 & 50.37 & 35.43 & \textbf{26.73}  & 33.46 & 25.05 & 26.02 & \textbf{76.00} & \textbf{89.47} & 46.76 & 4.00 & \textbf{95.50} & 60.79 & \textbf{62.51} & \textbf{46.97}\\

\cline{2-19}
& \multicolumn{18}{c}{\cellcolor{lightgray!25} 
  \textit{Compression Ratio = 0.5}
} \\

&~~~ThinK* & 29.27 & 27.37 & 39.90 & 45.95 & 31.83 & 20.41 & 22.24 & 23.21 & 19.80 & 65.50 & 87.27 & 43.43 & \textbf{6.00} & 80.00 & 59.03 & 59.32 & 41.28 \\

&~~~Palu & 25.84 & 35.28 & 47.43 & 46.69 & 31.93 & \textbf{27.21} & 29.32 & 24.32 & 25.37 & 74.50 & 86.31 & 42.95 & 4.50 & 61.00 & 48.19 & 47.90 & 41.17 \\

&~~~CommonKV & 27.27 & 34.71 & 46.70 & 44.74 & 27.02 & 24.21 & 26.36 & 23.52 & 24.42 & 51.00 & 88.56 & 43.82 & 4.50 & 88.00 & 60.14 & 59.94 & 42.18 \\

&~~~EchoKV & 30.26 & 36.92 & \textbf{49.71} & \textbf{51.33} & 34.26 & 24.51 & 31.54 & \textbf{25.28} & 25.36 & \textbf{76.00} & 89.32 & \textbf{46.93} & 3.50 & \textbf{93.50} & \textbf{61.72} & 61.66 & 46.36 \\

&~~~EchoKV-Hybrid & \textbf{31.32} & \textbf{38.69} & 49.55 & 49.05 & \textbf{34.95} & 25.10 & \textbf{32.88} & 25.02 & \textbf{25.83} & \textbf{76.00} & \textbf{89.57} & 46.55 & 3.50 & 93.00 & 61.00 & \textbf{62.78} & \textbf{46.55} \\

\cline{2-19}
& \multicolumn{18}{c}{\cellcolor{lightgray!25} 
  \textit{Compression Ratio = 0.3}
} \\

&~~~Palu & 11.04 & 11.91 & 23.13 & 16.00 & 14.48 & 8.77 & 10.60 & 20.50 & 16.87 & 58.50 & 58.98 & 26.55 & \textbf{4.50} & 4.08 & 23.34 & 25.56 & 20.93 \\

&~~~CommonKV & 11.96 & 21.25 & 29.05 & 21.47 & 14.49 & 8.40 & 22.30 & 22.55 & \textbf{24.43} & 56.50 & 72.81 & 37.21 & 2.63 & 14.50 & 52.46 & 39.16 & 28.20 \\

&~~~EchoKV & 27.99 & 35.01 & \textbf{47.29} & \textbf{49.77} & 31.63 & \textbf{24.01} & 25.88 & 23.86 & 23.63 & 67.00 & 89.06 & \textbf{45.00} & 4.00 & 70.00 & \textbf{59.27} & 61.05 & 42.78 \\

&~~~EchoKV-Hybrid & \textbf{28.01} & \textbf{37.11} & 47.08 & 47.98 & \textbf{33.35} & 23.63 & \textbf{27.11} & \textbf{23.95} & 22.93 & \textbf{75.50} & \textbf{89.71} & 44.41 & \textbf{4.50} & \textbf{72.00} & 58.96 & \textbf{61.62} & \textbf{43.62} \\

\midrule

\multirow{19}{*}{\rotatebox[origin=c]{90}{\fontsize{18}{100}\selectfont \textbf{Mistral-Small-24B-Instruct-2501}}}

&~~~FullKV & 33.72 & 47.19 & 54.95 & 67.24 & 64.93 & 49.04 & 33.22 & 24.54 & 25.47 & 75.50 & 93.71 & 48.86 & 21.00 & 100.0 & 63.67 & 70.66 & 54.61 \\
\cline{2-19}

& \multicolumn{18}{c}{\cellcolor{lightgray!25}
  \textit{Compression Ratio = 0.7}
} \\

&~~~MiniCache & 31.70 & 35.78 & 44.52 & 61.25 & 53.13 & 39.06 & 25.38 & 22.30 & 22.92 & 72.00 & 92.39 & 37.20 & 17.50 & 67.50 & 37.97 & 48.19 & 44.30 \\

&~~~ThinK & 34.43 & 41.31 & 51.85 & 67.44 & 64.09 & 47.23 & 24.61 & 23.98 & 20.62 & 61.00 & 93.24 & 46.86 & 19.00 & 100.0 & 61.30 & 69.05 & 51.63 \\

&~~~Palu & 34.71 & \textbf{47.09} & 55.35 & 66.61 & 63.78 & 44.51 & \textbf{32.68} & \textbf{25.72} & 25.53 & 74.50 & 92.12 & 48.22 & 14.50 & 100.0 & 54.65 & 65.63 & 52.85 \\

&~~~CommonKV & \textbf{36.16} & 46.59 & 54.82 & 66.00 & 62.80 & 46.29 & 32.13 & 24.85 & \textbf{25.55} & \textbf{76.00} & 93.16 & 48.20 & 18.00 & 99.50 & \textbf{68.02} & \textbf{71.11} & 54.32 \\

&~~~EchoKV & 33.74 & 46.73 & \textbf{55.47} & 67.18 & 64.07 & \textbf{49.50} & 31.71 & 25.28 & 24.92 & 75.50 & 93.64 & 48.22 & 19.50 & 100.0 & 64.38 & 70.98 & 54.43 \\

&~~~EchoKV-Hybrid & 33.72 & 47.04 & 55.20 & \textbf{67.46} & \textbf{64.89} & 49.09 & 32.06 & 25.54 & 25.20 & 75.50 & \textbf{93.71} & \textbf{48.55} & \textbf{21.00} & 100.0 & 63.33 & 71.04 & \textbf{54.58} \\

\cline{2-19}
& \multicolumn{18}{c}{\cellcolor{lightgray!25}
  \textit{Compression Ratio = 0.5}
} \\

&~~~ThinK* & 26.96 & 30.79 & 35.81 & 56.43 & 57.52 & 39.26 & 19.81 & 21.56 & 16.84 & 37.00 & 91.51 & 44.48 & 17.50 & 99.00 & 56.30 & 61.10 & 44.49 \\

&~~~Palu & 28.82 & 44.48 & 47.35 & 61.33 & 53.18 & 37.76 & 23.24 & 24.79 & 22.87 & 73.00 & 91.12 & 43.68 & 9.50 & 97.00 & 22.62 & 38.07 & 44.93 \\

&~~~CommonKV & \textbf{35.98} & 46.14 & 54.49 & 66.59 & 63.45 & 47.10 & 30.78 & 24.58 & \textbf{25.52} & 74.00 & 92.91 & 47.59 & 19.00 & 100.0 & \textbf{66.90} & \textbf{70.02} & 54.07 \\

&~~~EchoKV & 32.78 & \textbf{46.33} & 56.23 & \textbf{67.14} & \textbf{64.25} & 49.07 & \textbf{30.88} & 24.69 & 24.38 & \textbf{75.50} & 93.39 & \textbf{48.37} & \textbf{20.00} & 100.0 & 64.14 & 69.87 & \textbf{54.19} \\

&~~~EchoKV-Hybrid & 33.62 & 45.36 & \textbf{56.38} & 65.84 & 63.51 & \textbf{49.45} & 29.68 & \textbf{25.17} & 23.70 & \textbf{75.50} & \textbf{93.81} & 48.29 & \textbf{20.00} & 100.0 & 63.45 & 69.72 & 53.97 \\

\cline{2-19}
& \multicolumn{18}{c}{\cellcolor{lightgray!25}
  \textit{Compression Ratio = 0.3}
} \\

&~~~Palu & 4.77 & 6.85 & 16.03 & 7.47 & 5.98 & 6.47 & 7.06 & 15.09 & 10.45 & 36.50 & 22.94 & 16.84 & 1.00 & 0.00 & 21.35 & 22.41 & 12.58 \\

&~~~CommonKV & 27.41 & 41.87 & 48.97 & 55.45 & 53.02 & 28.62 & 26.16 & 22.74 & \textbf{24.72} & 72.50 & 90.17 & 45.51 & 1.23 & 95.00 & \textbf{70.49} & 68.78 & 48.29 \\

&~~~EchoKV & 30.77 & \textbf{44.80} & \textbf{54.43} & \textbf{66.78} & \textbf{62.74} & \textbf{49.01} & \textbf{27.57} & \textbf{24.66} & 23.05 & \textbf{74.50} & 93.23 & \textbf{45.76} & 18.50 & \textbf{99.50} & 63.25 & \textbf{69.95} & \textbf{53.03} \\

&~~~EchoKV-Hybrid & \textbf{31.21} & 34.22 & 43.76 & 62.49 & 58.14 & 41.53 & 20.42 & 22.87 & 17.58 & 41.00 & \textbf{93.33} & 45.41 & \textbf{21.00} & 99.00 & 58.13 & 62.71 & 47.05 \\

\bottomrule[2pt]
\end{tabular}
}
\end{threeparttable}
\end{table*}

\begin{figure}[t]
    \centering
    \includegraphics[width=0.95\linewidth]{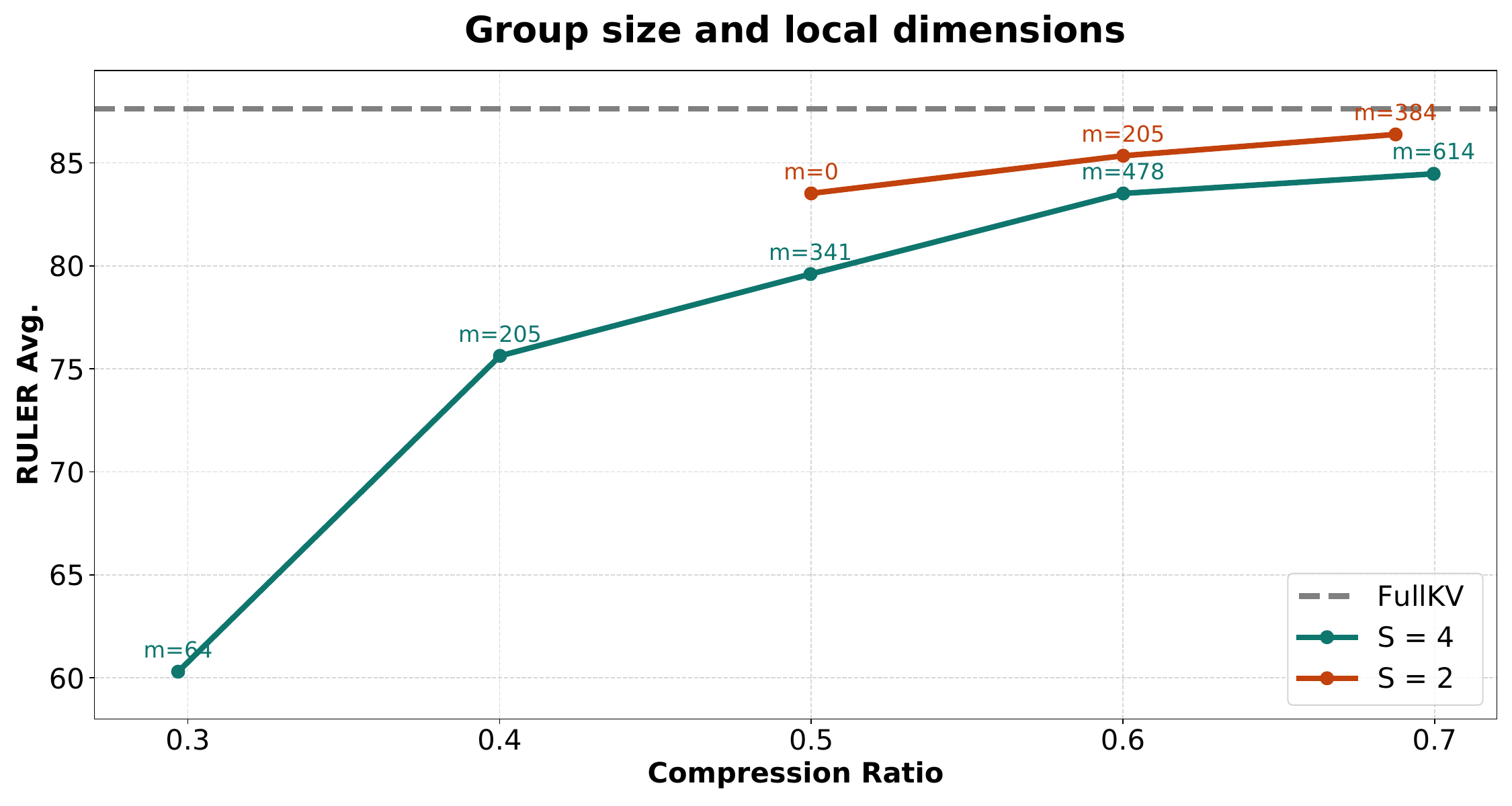}
    \caption{RULER average for different group sizes $S$ and local dimensions $m$ on Llama-3.1-8B-Instruct. The gray dashed line denotes the FullKV baseline, and each point is annotated with its retained local dimension $m$. Smaller $S$ corresponds to a shorter prediction distance to the retained full layer.}
    \label{fig:sm_ablation}
\end{figure}

\subsection{Visualization of Needle-in-a-Haystack}
\label{app:niah}

\begin{figure*}[t]
    \centering
    \begin{subfigure}[b]{\linewidth}
        \centering
        \includegraphics[width=\linewidth]{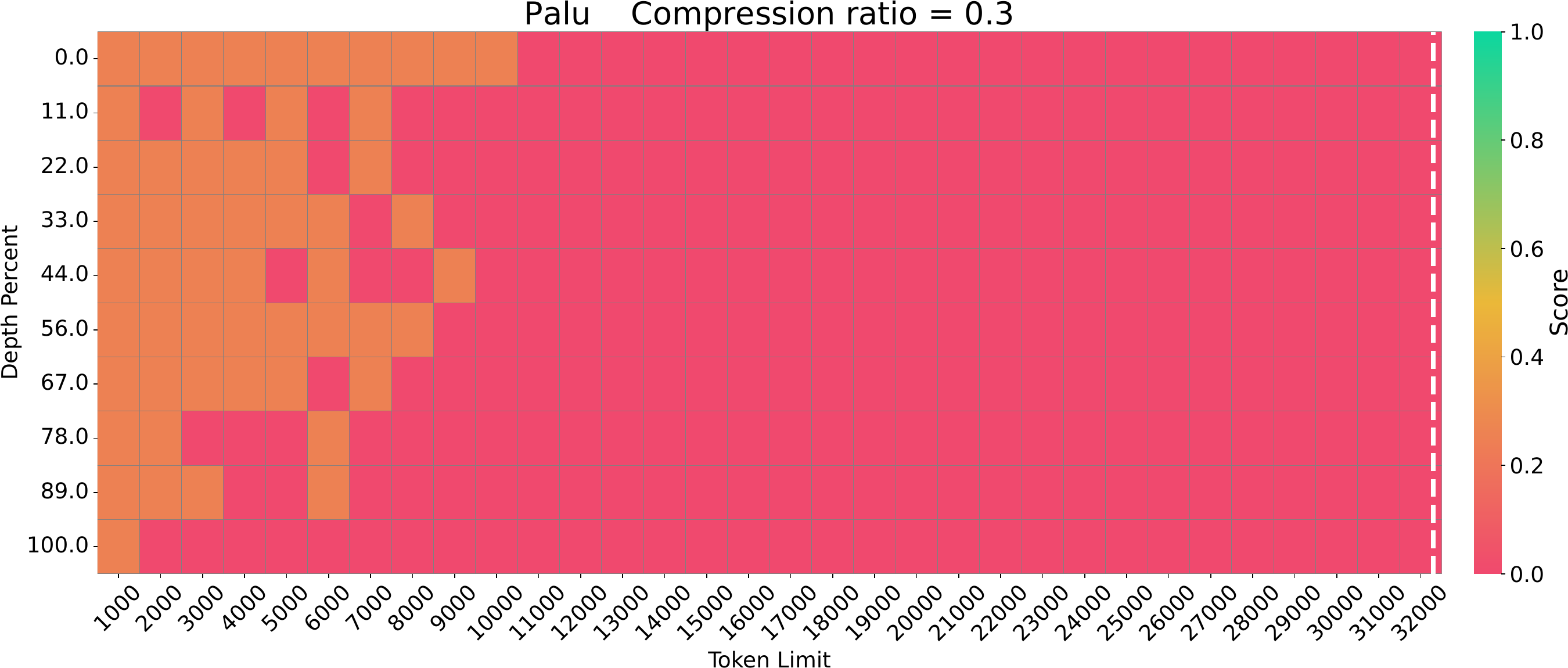}
        \caption{Palu (Score = 4.3)}
        \label{fig:llama_palu}
    \end{subfigure}
    
    \vspace{5pt} 
    
    \begin{subfigure}[b]{\linewidth}
        \centering
        \includegraphics[width=\linewidth]{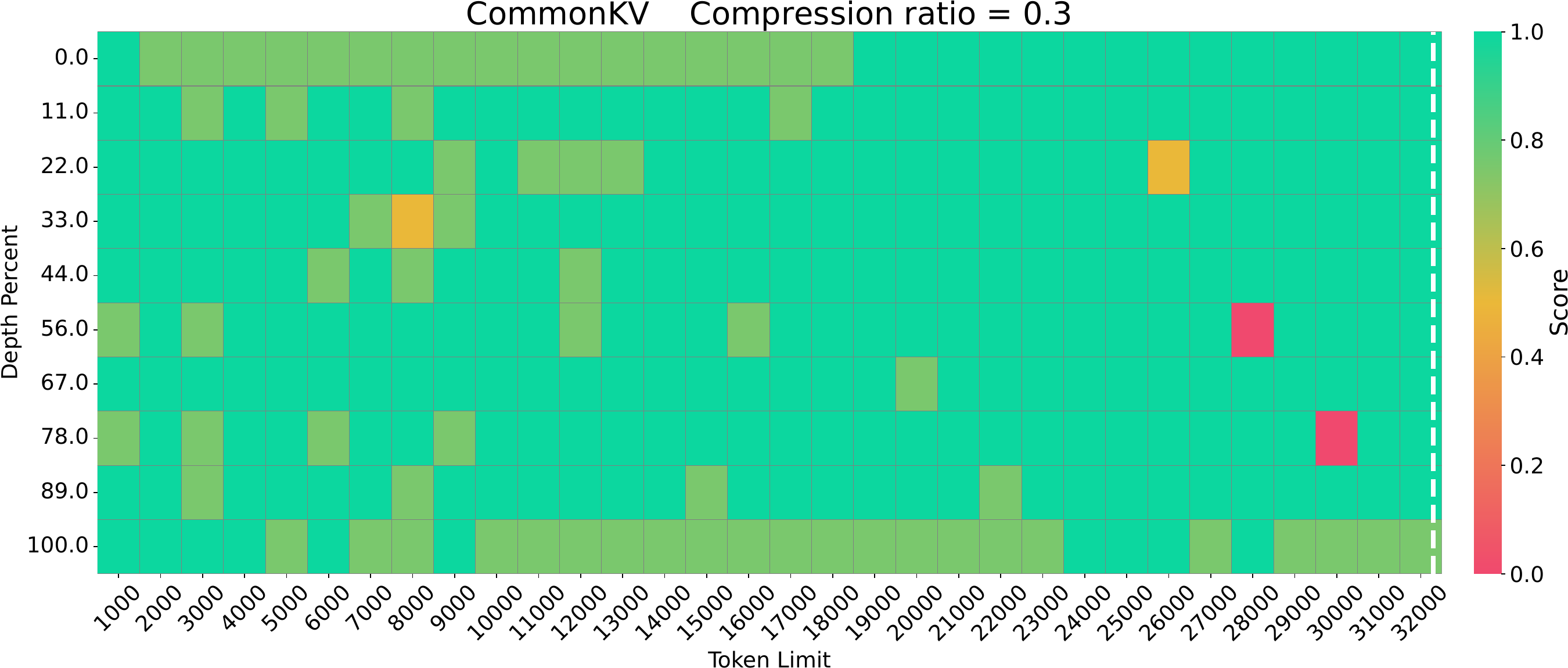}
        \caption{CommonKV (Score = 94.0)}
        \label{fig:llama_commonkv}
    \end{subfigure}
    
    \vspace{5pt}
    
    \begin{subfigure}[b]{\linewidth}
        \centering
        \includegraphics[width=\linewidth]{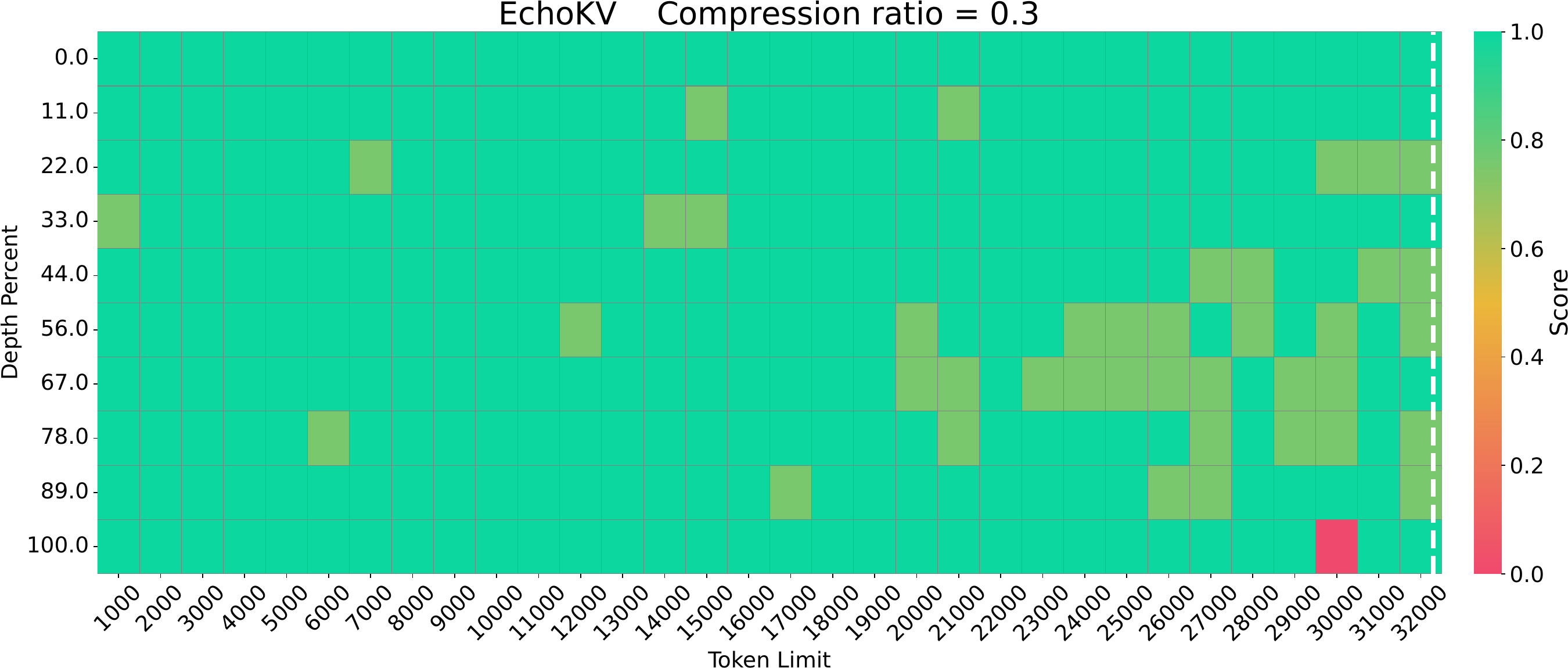}
        \caption{EchoKV (Score = 96.6)}
        \label{fig:llama_echo}
    \end{subfigure}
    \caption{Visualization of NIAH results on Llama-3.1-8B-Instruct with a compression ratio of 0.3.}
    \label{fig:llama_niah}
\end{figure*}

\begin{figure*}[t]
    \centering
    \begin{subfigure}[b]{\linewidth}
        \centering
        \includegraphics[width=\linewidth]{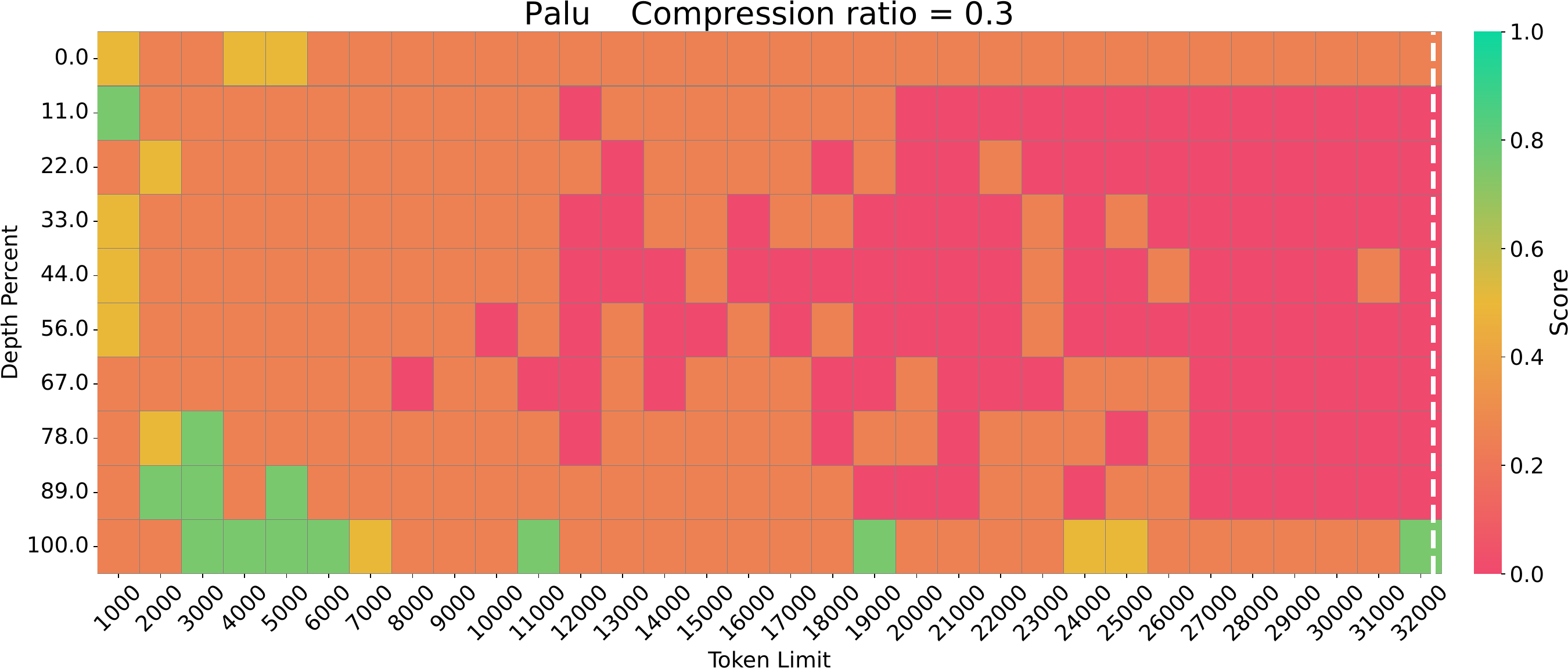}
        \caption{Palu (Score = 18.9)}
        \label{fig:mistral_palu}
    \end{subfigure}
    
    \vspace{5pt}
    
    \begin{subfigure}[b]{\linewidth}
        \centering
        \includegraphics[width=\linewidth]{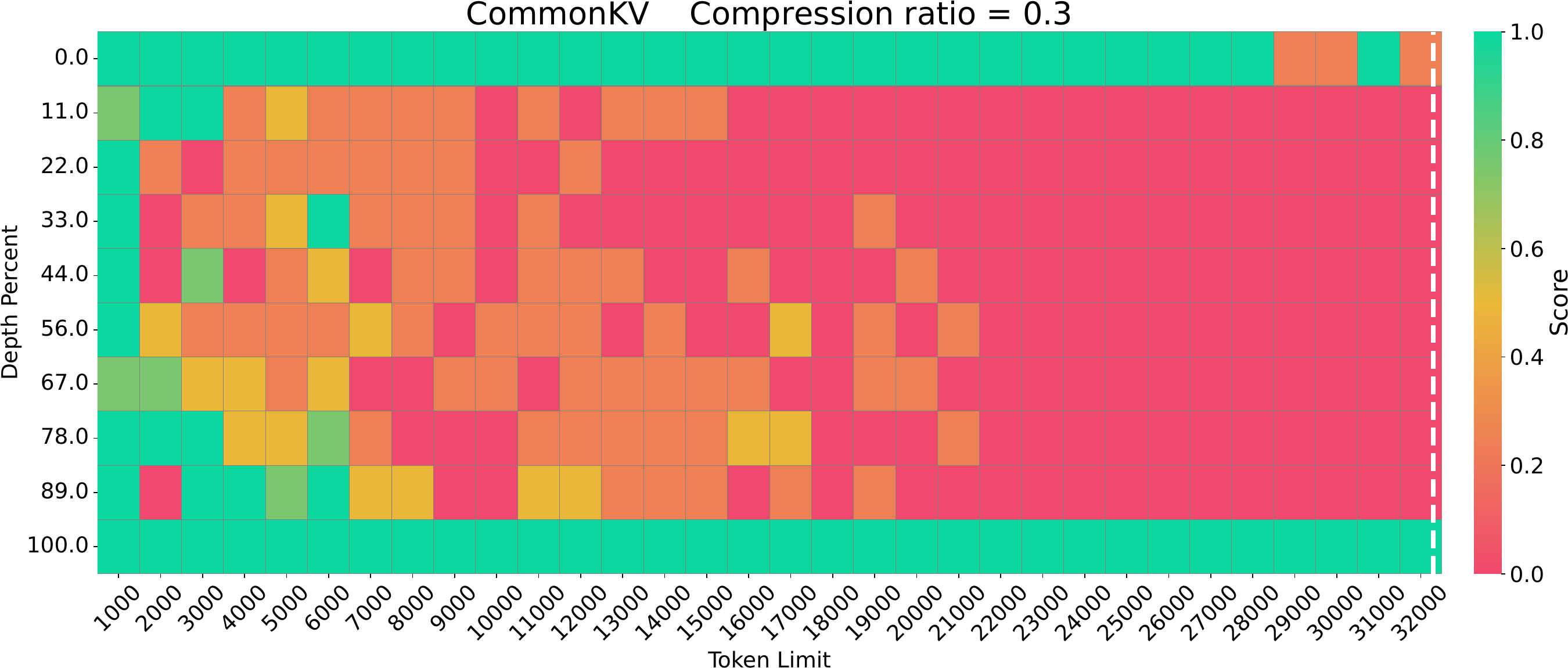}
        \caption{CommonKV (Score = 32.8)}
        \label{fig:mistral_commonkv}
    \end{subfigure}
    
    \vspace{5pt}
    
    \begin{subfigure}[b]{\linewidth}
        \centering
        \includegraphics[width=\linewidth]{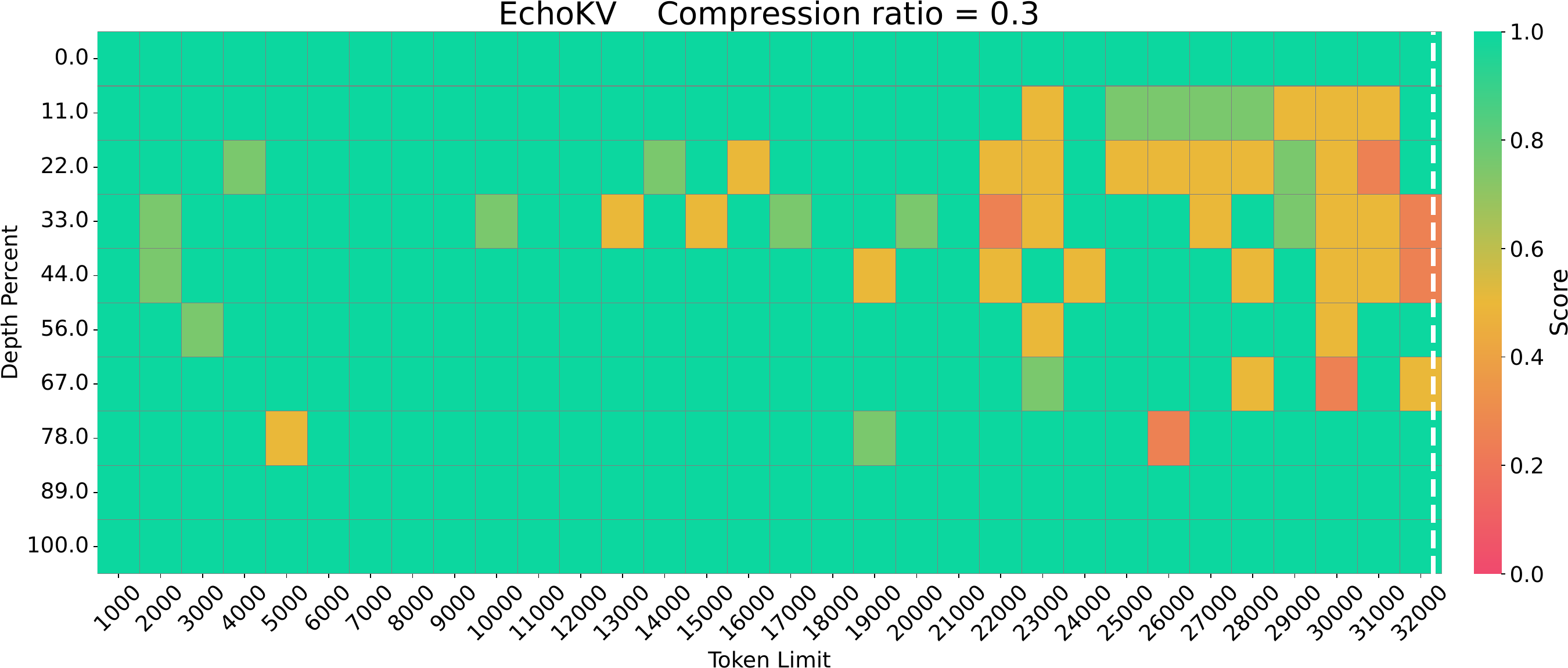}
        \caption{EchoKV (Score = 92.8)}
        \label{fig:mistral_echo}
    \end{subfigure}
    \caption{Visualization of NIAH results on Mistral-7B-Instruct-v0.3 with a compression ratio of 0.3.}
    \label{fig:mistral_niah}
\end{figure*}

To provide a more intuitive comparison of the retrieval capabilities of different methods under low memory budgets, we visualize the results of the ``Needle-in-a-Haystack'' (NIAH) task.
Figure \ref{fig:llama_niah} and Figure \ref{fig:mistral_niah} illustrate the performance of Palu, CommonKV, and EchoKV on Llama-3.1-8B-Instruct and Mistral-7B-Instruct-v0.3, respectively, at a compression ratio of 0.3.
Green regions indicate successful retrieval, while yellow and red regions denote failures.
As observed, EchoKV maintains consistently superior coverage across the context window on both models, demonstrating strong generalization capabilities.
In contrast, while CommonKV performs relatively well on Llama-3.1, it suffers from severe information loss on Mistral-7B, further highlighting the robustness of our approach across different models.

\subsection{Adaptive inference and switching}
\label{app:switching_results}

Table \ref{tab:adaptive_workload} evaluates the policies in Appendix \ref{app:inference}.
Both adaptive policies complete all 20 batches while retaining FullKV-like TPOT for short prompts.
The memory policy switches during prefill at layers 19/14 for 32K/36K, costing 193.52/165.52 ms; the length policy selects EchoKV before prefill.

\begin{table}[htbp]
    \centering
    \caption{Adaptive workload on Llama-3.1-8B-Instruct, Pre-RoPE, nominal ratio 0.5, batch size 8, and 16 output tokens. Latencies are averages over five batches per length (20 batches total). Short/long prompts correspond to 4K/8K and 32K/36K. $^*$The evaluated official Palu implementation stores reconstructed full-dimensional K/V in the standard attention cache.}
    \label{tab:adaptive_workload}
    \footnotesize
    \setlength{\tabcolsep}{2pt}
    \renewcommand{\arraystretch}{1.15}
    \newcommand{\workloadcell}[2][c]{\begin{tabular}[c]{@{}#1@{}}#2\end{tabular}}
    \begin{tabular*}{\linewidth}{@{\extracolsep{\fill}}lcccccc@{}}
    \toprule
    \textbf{Policy} & \workloadcell{\textbf{Short TPOT (ms)}\\\textbf{4K/8K}} & \workloadcell{\textbf{Long TPOT (ms)}\\\textbf{32K/36K}} & \workloadcell{\textbf{Routing}\\\textbf{short / long}} & \workloadcell{\textbf{Switch (ms)}\\\textbf{32K/36K}} & \workloadcell{\textbf{KV cache (GiB)}\\\textbf{short; long}} & \workloadcell{\textbf{Completed}\\\textbf{batches}} \\
    \midrule
    FullKV-only & 36.33 / 50.52 & OOM / OOM & FullKV only & - & \workloadcell{4.015 / 8.015;\\OOM / OOM} & 10/20 \\
    \addlinespace[3pt]
    Echo static & 74.27 / 129.65 & 469.54 / 528.20 & EchoKV only & - & \workloadcell{2.079 / 4.079;\\16.079 / 18.079} & 20/20 \\
    \addlinespace[3pt]
    \workloadcell[l]{Echo-adaptive\\(length)} & 35.00 / 50.16 & 472.67 / 530.62 & FullKV / EchoKV & - & \workloadcell{4.015 / 8.015;\\16.079 / 18.079} & 20/20 \\
    \addlinespace[3pt]
    \workloadcell[l]{Echo-adaptive\\(memory)} & 36.08 / 50.19 & 475.72 / 534.09 & \workloadcell{FullKV /\\FullKV$\to$EchoKV} & 193.52 / 165.52 & \workloadcell{4.015 / 8.015;\\16.079 / 18.079} & 20/20 \\
    \addlinespace[3pt]
    Palu static$^*$ & 43.71 / 51.47 & OOM / OOM & Palu only & - & \workloadcell{4.015 / 8.015;\\OOM / OOM} & 10/20 \\
    \bottomrule
    \end{tabular*}
\end{table}

Table \ref{tab:switch_quality} reports quality after switching during generation.
Switch@32/64 yields macro averages of 26.63--27.09, versus 26.18 for static EchoKV and 28.99 for FullKV; later switching retains more of the FullKV trajectory.

\begin{table}[htbp]
    \centering
    \caption{LongBench summarization results for Llama-3.1-8B-Instruct at ratio 0.5. Static EchoKV stores the compressed cache from prefill; Switch@$t$ starts compression after $t$ generated tokens.}
    \label{tab:switch_quality}
    \small
    \begin{tabular}{lrrrr}
    \toprule
    \textbf{Mode} & \textbf{GovReport} & \textbf{QMSum} & \textbf{MultiNews} & \textbf{Average} \\
    \midrule
    FullKV & 34.66 & 25.19 & 27.11 & 28.99 \\
    Static EchoKV & 29.22 & 24.10 & 25.21 & 26.18 \\
    Switch@32, Pre-RoPE & 29.72 & 24.75 & 25.54 & 26.67 \\
    Switch@32, Post-RoPE & 29.78 & 24.58 & 25.52 & 26.63 \\
    Switch@64, Pre-RoPE & 30.34 & 25.27 & 25.67 & 27.09 \\
    Switch@64, Post-RoPE & 30.16 & 25.07 & 25.71 & 26.98 \\
    \bottomrule
    \end{tabular}
\end{table}

In the separate batch-1 Switch@32 test, converting 16K/32K caches takes 24.04/45.97 ms with Pre-RoPE and 3.21/3.53 ms with Post-RoPE.

\subsection{Ultra-long context}
\label{app:ultralong}

We evaluate Llama-3.1-8B-Instruct at 64K and 128K on all 13 RULER tasks, using greedy decoding and macro averaging.
EchoKV and Palu use ratio 0.5; ThinK uses an overall KV ratio of 0.6.
As shown in Table \ref{tab:ultralong}, EchoKV exceeds Palu by 17.32/26.24 points at 64K/128K, with gaps to FullKV of 4.50/12.92 points.
EchoKV completes the 128K evaluation, where ThinK runs out of memory.

\begin{table}[htbp]
    \centering
    \caption{RULER macro averages at 64K and 128K on Llama-3.1-8B-Instruct. Ratios are shown in parentheses.}
    \label{tab:ultralong}
    \small
    \begin{tabular}{lrrrr}
    \toprule
    \textbf{Context} & \textbf{FullKV} & \textbf{Palu (0.5)} & \textbf{ThinK (0.6)} & \textbf{EchoKV (0.5)} \\
    \midrule
    64K & 84.13 & 62.31 & 10.77 & \textbf{79.63} \\
    128K & 75.46 & 36.30 & OOM & \textbf{62.54} \\
    \bottomrule
    \end{tabular}
\end{table}

\subsection{Seed robustness}
\label{app:seed_robustness}

We independently train Llama-3.1-8B-Instruct predictors with seeds 42--45 at ratios 0.7 and 0.5.
Table \ref{tab:seed_robustness} reports means and sample standard deviations over these four predictors; greedy evaluation is deterministic for a fixed predictor.
The standard deviations of the LongBench and 32K RULER averages are at most 0.14 and 0.27, respectively.
Individual LongBench categories vary more, particularly synthetic tasks at ratio 0.7.

\begin{table}[htbp]
    \centering
    \caption{Predictor-training seed variation on Llama-3.1-8B-Instruct. EchoKV entries are mean $\pm$ sample standard deviation over seeds 42--45; FullKV is deterministic.}
    \label{tab:seed_robustness}
    \small
    \begin{tabular}{lccc}
    \toprule
    \textbf{Metric} & \textbf{FullKV} & \textbf{EchoKV (0.7)} & \textbf{EchoKV (0.5)} \\
    \midrule
    Single-document QA & 44.94 & $45.02\pm0.15$ & $44.81\pm0.21$ \\
    Multi-document QA & 46.88 & $46.80\pm0.12$ & $46.75\pm0.07$ \\
    Summarization & 28.99 & $27.39\pm0.62$ & $26.26\pm0.12$ \\
    Few-shot learning & 69.60 & $69.20\pm0.12$ & $68.82\pm0.17$ \\
    Synthetic & 53.28 & $50.02\pm2.10$ & $47.84\pm0.35$ \\
    Code & 60.14 & $59.91\pm0.54$ & $59.83\pm0.22$ \\
    \midrule
    LongBench average & 49.88 & $49.07\pm0.14$ & $48.45\pm0.08$ \\
    RULER average & 87.63 & $86.27\pm0.10$ & $83.50\pm0.27$ \\
    \bottomrule
    \end{tabular}
\end{table}

\section{Additional analysis}
\label{app:additional_analysis}
\subsection{Analysis of QK-KL divergence loss}
\label{app:qk_loss}

In this section, we provide the detailed formulation of the standard KL divergence loss (QK-KL) referenced in \S\ref{sec:train}, and analyze the computational bottlenecks that led us to adopt the Output MSE loss instead.

\paragraph{Formulation.}
The primary objective of the QK-KL loss is to ensure that the attention probability distribution generated by the reconstructed Key cache closely approximates the ground truth distribution.
Given a query matrix $\mathbf{Q} \in \mathbb{R}^{L \times d}$ and the original Key matrix $\mathbf{K} \in \mathbb{R}^{L \times d}$, the standard attention weights $\mathbf{A}$ are computed via the softmax operation:
\begin{equation}
    \mathbf{A} = \text{softmax}\left( \frac{\mathbf{Q}\mathbf{K}^T}{\sqrt{d}} \right),
\end{equation}
where $\mathbf{A} \in \mathbb{R}^{L \times L}$ represents the full attention matrix, and $d$ denotes the head dimension.

Similarly, utilizing the reconstructed Key matrix $\tilde{\mathbf{K}}$, which is composed of the retained local cache and the predicted dropped cache (as defined in Eq. \ref{eq:concat}), we obtain the approximated attention weights $\tilde{\mathbf{A}}$:
\begin{equation}
    \tilde{\mathbf{A}} = \text{softmax}\left( \frac{\mathbf{Q}\tilde{\mathbf{K}}^T}{\sqrt{d}} \right).
\end{equation}

The QK-KL loss is then defined as the Kullback-Leibler divergence between the target distribution $\mathbf{A}$ and the approximated distribution $\tilde{\mathbf{A}}$:
\begin{equation}
\begin{split}
    \mathcal{L}_{\text{KL}} &= \frac{1}{L} \sum_{i=1}^{L} D_{\text{KL}}(\mathbf{A}_i \parallel \tilde{\mathbf{A}}_i) \\
    &= \frac{1}{L} \sum_{i=1}^{L} \sum_{j=1}^{L} \mathbf{A}_{i,j} \log \left( \frac{\mathbf{A}_{i,j}}{\tilde{\mathbf{A}}_{i,j}} \right),
\end{split}
\end{equation}
where $\mathbf{A}_i$ and $\tilde{\mathbf{A}}_i$ represent the attention distribution for the $i$-th token.

\paragraph{Complexity analysis.}
While $\mathcal{L}_{\text{KL}}$ effectively aligns the attention distributions, it imposes significant computational and memory overhead in long-context scenarios.
Calculating the loss requires explicitly materializing the full $L \times L$ attention probability matrices $\mathbf{A}$ and $\tilde{\mathbf{A}}$.
This results in a quadratic memory complexity of $O(L^2)$, which negates the memory efficiency benefits of sparse KV caching.
Furthermore, this explicit materialization is incompatible with memory-efficient attention kernels such as FlashAttention \citep{dao2022flashattention}, which avoid storing the intermediate attention matrix to achieve linear memory complexity.
Consequently, we opt for the Output MSE loss (Eq. \ref{eq:o_loss}), which operates on the attention output $\mathbf{O} \in \mathbb{R}^{L \times d}$ and maintains compatibility with efficient kernels.

\subsection{Group size, local dimension, and prediction distance}
\label{app:sm_ablation}

We further study the interaction between the group size $S$ and the retained local dimension $m$.

\paragraph{Cache budget and construction.}
For a complete group of $S$ layers, each token in the compressed region stores $D=H_{\mathrm{KV}}d_{\mathrm{head}}$ features at the anchor and $m$ at each other layer.
The nominal cache ratio is
\begin{equation}
    r_{\mathrm{nom}}(S,m)=\frac{D+(S-1)m}{SD}.
    \label{eq:cache_ratio}
\end{equation}
K and V share this ratio under the same configuration and precision.
Anchor storage sets a lower bound of $1/S$.
For a target $0<r_{\mathrm{target}}<1$, we choose the smallest feasible group and allocate the remaining budget to local features:
\begin{equation}
    S=\left\lceil\frac{1}{r_{\mathrm{target}}}\right\rceil,
    \qquad
    m^{*}=\frac{D(Sr_{\mathrm{target}}-1)}{S-1}.
    \label{eq:budget_construction}
\end{equation}
The continuous budget $m^{*}$ is rounded to a supported dimension $m$; Appendix \ref{app:config} gives the configurations and resulting ratios.
FullKV corresponds to ratio 1.
Choosing a small $S$ keeps the anchor close to its target layers, while $m$ preserves same-layer information within the remaining budget.
Appendix \ref{app:inference} describes full-token storage and other memory components.

\paragraph{Empirical analysis.}
To examine this design choice, we evaluate multiple $(S, m)$ combinations on Llama-3.1-8B-Instruct and report their RULER averages in Figure \ref{fig:sm_ablation}.

Two trends are clear.
First, under comparable compression ratios, smaller group sizes tend to perform better.
For example, around a compression ratio of 0.5, the $S=2, m=0$ setting outperforms the nearby $S=4, m=341$ setting, and around 0.6, the $S=2, m=205$ setting also exceeds the corresponding $S=4, m=478$ setting.
This suggests that reducing the prediction distance between the target layer and the retained full layer improves reconstruction quality.
Second, when $S$ is fixed, increasing $m$ consistently improves performance, since more local information from the current layer is preserved.
At the same time, $S$ cannot be made arbitrarily small for all budgets: for instance, with $S=2$, the lowest feasible compression ratio is 0.5 when no local dimensions are retained.
Therefore, lower compression ratios require larger group sizes in order to satisfy the target memory budget.

This trade-off provides a simple interpretation of the $(S, m)$ design space.
The group size $S$ controls how far the predictor must extrapolate across layers, while the local dimension $m$ controls how much same-layer information is directly retained.
Smaller $S$ is therefore preferable whenever the memory budget allows it, and larger $m$ should be used within a fixed $S$ because it reduces the amount of information that must be reconstructed.

\subsection{Local-dimension selection}
\label{app:local_select}

We also compare different strategies for selecting the retained local dimensions.
Besides the default prefix strategy that keeps the first $m$ flattened dimensions, we evaluate three alternatives: random selection, a per-head allocation strategy that spreads the retained dimensions across heads, and an oracle-style greedy strategy based on similarity to the dropped dimensions.
We report results on Llama-3.1-8B-Instruct under two representative settings: a moderate-budget configuration $(S=2, m=384)$ with a compression ratio of 0.7, and a lower-budget configuration $(S=4, m=64)$ with a compression ratio of 0.3.

\begin{table*}[t]
    \centering
    \caption{Comparison of local-dimension selection strategies on Llama-3.1-8B-Instruct. Prefix denotes the default contiguous selection used by EchoKV.}
    \label{tab:local_select}
    \begin{tabular}{llcccc}
    \toprule
    \textbf{Ratio} & \textbf{Benchmark} & \textbf{Prefix} & \textbf{Random} & \textbf{Per-head} & \textbf{Oracle} \\
    \midrule
    \multirow{2}{*}{0.7 $(S=2, m=384)$}
    & LongBench & 49.075 & 49.509 & 49.177 & \textbf{49.553} \\
    & RULER     & \textbf{86.386} & 84.765 & 82.805 & 85.440 \\
    \midrule
    \multirow{2}{*}{0.3 $(S=4, m=64)$}
    & LongBench & 45.267 & 45.916 & \textbf{46.038} & 45.205 \\
    & RULER     & \textbf{60.302} & 48.446 & 43.125 & 60.159 \\
    \bottomrule
    \end{tabular}
\end{table*}

Table \ref{tab:local_select} shows that the simple prefix strategy is already highly competitive.
At the moderate-budget setting, more sophisticated strategies bring only marginal gains on LongBench and do not improve RULER, indicating that contiguous retention is sufficient when enough local information is preserved.
At the more aggressive setting, prefix selection remains the most robust choice on RULER and is close to the best LongBench result.
These results suggest that the main benefit comes from retaining a small amount of same-layer information, while the exact choice of dimensions is less critical than the overall budget.
Given this small performance gap, the prefix strategy remains attractive because it avoids additional selection overhead and preserves contiguous memory access.

\subsection{Prediction network architecture}
\label{app:pred_arch}

We further examine whether a more expressive predictor architecture is beneficial.
In addition to the default linear predictor, we evaluate two one-hidden-layer MLP variants on Llama-3.1-8B-Instruct at a compression ratio of 0.5: a parameter-matched MLP whose width is chosen to keep the parameter count comparable to the linear predictor, and a wider MLP whose hidden size equals the predictor input dimension.

\begin{table}[t]
    \centering
    \caption{Comparison of prediction network architectures on Llama-3.1-8B-Instruct at a compression ratio of 0.5.}
    \label{tab:pred_arch}
    \begin{tabular}{lcc}
    \toprule
    \textbf{Architecture} & \textbf{LongBench Avg.} & \textbf{RULER Avg.} \\
    \midrule
    Linear & 48.53 & 83.52 \\
    MLP (param-matched) & 48.80 & 82.70 \\
    MLP (hidden = input dim) & \textbf{48.84} & \textbf{83.94} \\
    \bottomrule
    \end{tabular}
\end{table}

Table \ref{tab:pred_arch} shows that increasing predictor complexity yields only modest gains.
On LongBench, both MLP variants remain within 0.31 average points of the linear baseline, and on RULER the best MLP improves the average by only 0.42 points.
Moreover, the parameter-matched MLP does not consistently outperform the linear model, indicating that additional nonlinearity alone is not sufficient to deliver reliable gains.
Given that larger predictors also introduce extra reconstruction overhead during inference, these results do not justify moving away from the simple linear architecture.
We therefore retain the linear predictor in all main experiments.

\section{Use of AI tools}
\label{app:ai}
We acknowledge the use of AI tools solely for the purpose of language polishing and grammatical refinement of this manuscript. All core methodologies, experimental designs, and data analyses are conducted independently
by the authors without the involvement of AI-generated content.

\end{document}